\PassOptionsToPackage{numbers, compress}{natbib}
\PassOptionsToPackage{table}{xcolor}

\documentclass{article}

\usepackage[preprint]{neurips_2026}

\usepackage{amsmath}
\usepackage[utf8]{inputenc} % allow utf-8 input
\usepackage[T1]{fontenc}    % use 8-bit T1 fonts
\usepackage{hyperref}       % hyperlinks
\usepackage{url}            % simple URL typesetting
\usepackage{booktabs}       % professional-quality tables
\usepackage{amsfonts}       % blackboard math symbols
\usepackage{nicefrac}       % compact symbols for 1/2, etc.
\usepackage{microtype}      % microtypography
\usepackage{xcolor}         % colors
\usepackage{comment}
\usepackage[export]{adjustbox}
\usepackage{stfloats}
\usepackage{afterpage}
\usepackage{tikz}
\usepackage{fontawesome5}
\usepackage{booktabs}
\usepackage{rotating}
\usepackage{makecell}
\usepackage{multirow}
\usepackage[table]{xcolor}
\usepackage{booktabs}
\usepackage{array}
\usepackage{pifont} 
\definecolor{highlight}{gray}{0.95}
\usepackage{enumitem}
\setlist[itemize]{leftmargin=1em, itemsep=1pt, topsep=1pt}

\title{PHMForge: Evaluating LLM Agents on Industrial Prognostics through MCP-Native, Algorithm-Grounded Tools}

\author{%
  Yusheng Li \\
  Columbia University \\
  \texttt{yl6009@columbia.edu} \\
  \And
  Tianjun Feng \\
  Columbia University \\
  \texttt{tf2637@columbia.edu} \\
  \And
  Yunfeng Chen \\
  Columbia University \\
  \texttt{yc4640@columbia.edu} \\
  \AND
  Chun-Yi Tsai \\
  Columbia University \\
  \texttt{ct3316@columbia.edu} \\
  \And
  Yihan Sun \\
  Columbia University \\
  \texttt{ys3976@columbia.edu} \\
  \And
  Ayan Das \\
  Georgia Institute of Technology \\
  \texttt{adas446@gatech.edu} \\
  \And
  Kaoutar El Maghraoui \\
  IBM, New York \\
  \texttt{kelmaghr@us.ibm.com} \\
  \And
  Shuxin Lin \\
  IBM, New York \\
  \texttt{shuxin.lin@ibm.com} \\
  \And
  Dhaval Patel \\
  IBM, New York \\
  \texttt{pateldha@us.ibm.com} \\
}
 
\begin{document}
\maketitle

% =====================================================================
%  PHMForge — NeurIPS 2026 Datasets & Benchmarks Track submission
%  Rewritten in MCPMark structural style
%  (Page-1 pipeline figure; comparison table in §1; combined §2;
%   failure decomposition as figure; pass-all-3 consistency metric.)
% =====================================================================

% ---------------------------------------------------------------------
% ABSTRACT
% Open with the deployment-substrate framing (MCP at industrial scale),
% then the three-conflations methodology hook, then the headline
% numbers in the second half. Mirrors MCPMark's abstract structure.
% ---------------------------------------------------------------------
\begin{abstract}
LLM agents are beginning to invoke industrial asset-management tools
through the Model Context Protocol (MCP), yet whether they can act
reliably on this substrate for safety-critical \emph{Prognostics and
Health Management (PHM)} is unanswered. Prior benchmarks conflate
protocol fluency with reasoning, instrumentation failures with agent
failures, and tool use with tool retrieval. We introduce
\textbf{PHMForge}, an evaluation environment that closes each
conflation. PHMForge ships 99 SME-authored scenarios across eight 
industrial asset classes spanning rotating equipment, aero-engines, 
and lithium-ion battery cells, on real public datasets including 
NASA PCoE. The benchmark is served through 39 MCP-native tools that 
wrap published PHM algorithms (e.g., C-MAPSS, ISO~10816, Arrhenius 
capacity-fade models, and time-series foundation models for 
sequence forecasting). Krippendorff's $\alpha \in [0.74,\,0.82]$ 
on a 30-scenario stratified rotating-equipment/aero-engine sample; 
the lithium-ion battery extension is single-rater 
(Appendix~\ref{app:iaa-bess}). Across 
three agentic frameworks and six LLM backbones, the strongest 
configuration reaches \textbf{80.8\% pass@1} on the full 99-scenario 
set, with the residual gap concentrated in orchestration and 
tool-sequencing errors. Crucially, an architectural ablation 
reveals that replacing MCP tool execution with text-based 
Retrieval-Augmented Generation (RAG) over telemetry-equivalent 
evidence collapses Remaining Useful Life (RUL) prediction 
\emph{pass-all-3} from \textbf{100\% to 20\%} (5/5 vs.\ 1/5 
scenarios) on the lithium-ion battery class, exposing the 
structural limits of static retrieval for prognostic 
computation. Trajectory-level decomposition shows orchestration 
errors dominate failures across backbones, while schema-invalid 
tool calls are concentrated in smaller open-weight models and 
rare in frontier configurations. Frontier LLMs are stronger at 
calling tools than at planning when to call them. PHMForge is 
open-sourced with deterministic evaluators, a public leaderboard, 
and a datasheet. A full-suite evaluation costs approximately 
\$20--\$50 in API spend, depending on backbone.
\end{abstract}
% =====================================================================
% SECTION 1 : INTRODUCTION
% MCPMark layout: hook → thesis → page-1 pipeline figure → three-
% conflations as named items → comparison table at end → contributions
% → headline preview.
% =====================================================================
\section{Introduction}
\label{sec:intro}

\begin{figure*}[t]
    \centering
    \includegraphics[width=1.03\textwidth]{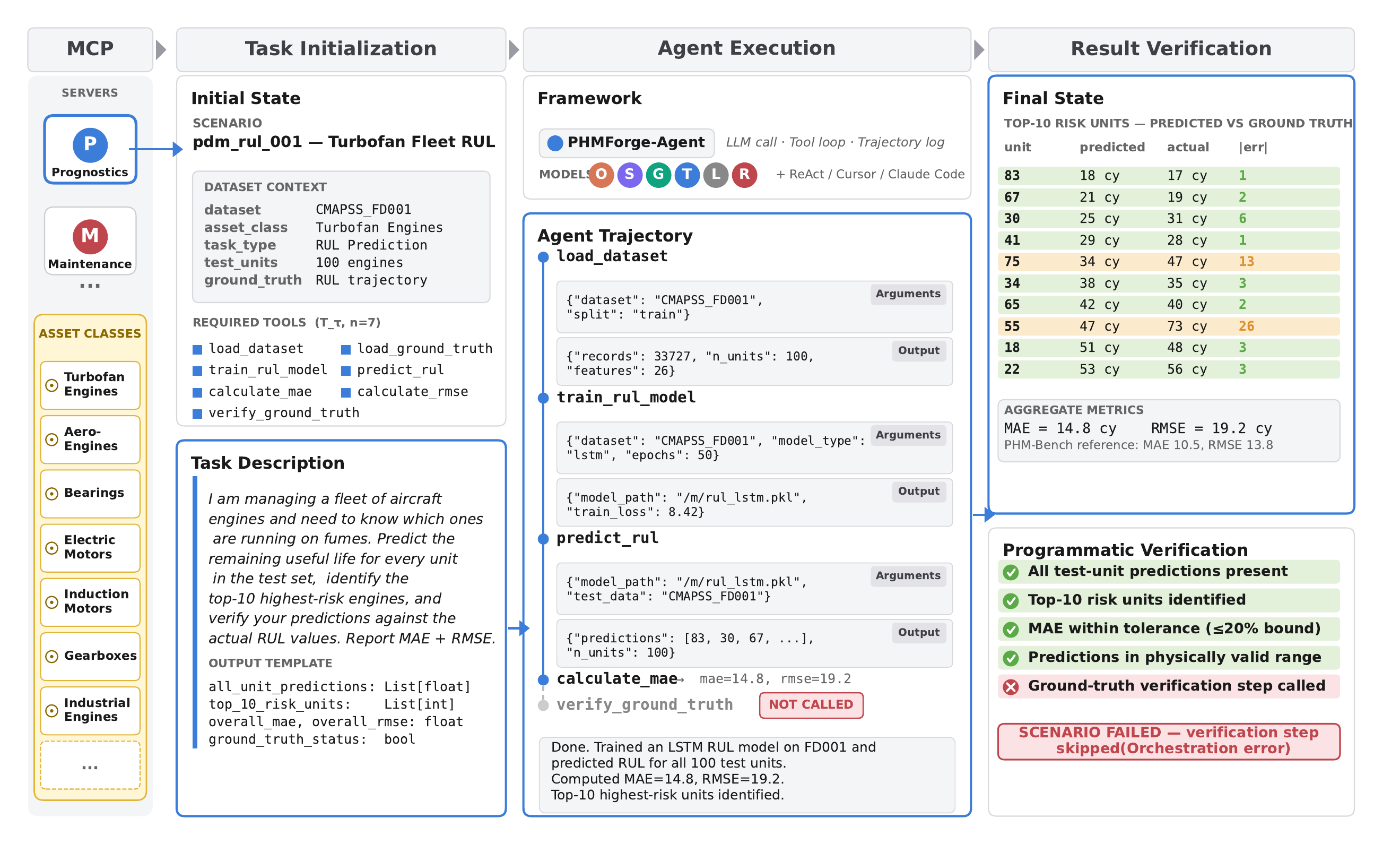}
    \caption{\textbf{PHMForge evaluation pipeline with full state tracking.} Each scenario begins from an SME-authored industrial query and a curated dataset context. The PHMForge-Agent (e.g., ReAct, ReActXen) then executes a tool-calling loop over algorithm-grounded MCP tools, followed by a deterministic verifier that evaluates whether all required checks are satisfied.}
    \label{fig:pipeline}
    \vspace{-0.25in}
\end{figure*}

When a turbofan engine sensor flags an anomaly mid-flight, or a 
wind-farm gearbox vibrates outside its ISO~10816 
envelope~\cite{iso10816}, the cost of a wrong decision is 
measured in millions of dollars, environmental damage, or human 
lives. \textbf{Industrial Artificial 
Intelligence}~\cite{9378369} addresses this regime where failure 
is physical rather than digital, and its central discipline, 
\textbf{Prognostics and Health Management (PHM)}~\cite{PHMbook}, 
governs the lifecycle of critical assets from turbofan engines 
to industrial gearboxes. \textit{This paper argues that 
evaluating LLM agents for industrial PHM requires evaluation 
methodology that current benchmarks do not provide, and shows 
what that methodology should look like.} 
Figure~\ref{fig:pipeline} previews the evaluation pipeline on a 
worked turbofan-RUL scenario.

% --- Page-1 pipeline figure (MCPMark Figure 1 equivalent) ---

\textbf{Large Language Model agents}, paired with reasoning 
frameworks such as 
\textbf{ReAct}~\cite{yao2023reactsynergizingreasoningacting}, 
promise to break the bespoke-pipeline bottleneck of classical 
PHM, invoking tools served through the \textbf{Model Context 
Protocol (MCP)}~\cite{mcp_intro_2024}. In December~2025, 
Anthropic transferred MCP to the Linux Foundation's Agentic AI 
Foundation, now counting more than 10{,}000 published 
servers~\cite{linuxfoundation_aaif}, and the first 
asset-management integrations have appeared (e.g., Maximo MCP 
server~\cite{maximo_mcp_marketplace}). Whether current LLM 
agents are reliable enough to act on this substrate is an open 
question that existing evaluation cannot answer.

Existing agent benchmarks each illuminate one face of the problem 
and miss the rest. PHM-specific 
benchmarks (PDMBench~\cite{zhang2026pdmbench}, 
PHM-Bench~\cite{yang2025phmbenchdomainspecificbenchmarkingframework}, 
ITFormer/EngineMT-QA~\cite{wang2025itformer}) measure prediction or 
QA accuracy on fixed datasets, treating models as passive 
predictors rather than active orchestrators. Generic agent 
benchmarks (MLE-Bench~\cite{chan2025mlebenchevaluatingmachinelearning}, 
MCP-Universe~\cite{luo2025mcpuniversebenchmarkinglargelanguage}, 
StableToolBench~\cite{guo2025stabletoolbenchstablelargescalebenchmarking}) 
evaluate multi-step tool reasoning over generic digital domains; a 
parallel safety-focused line (MCPTox~\cite{wang2026mcptox}, 
MCP-SafetyBench~\cite{zong2026mcpsafetybenchbenchmarksafetyevaluation}, 
MCPMark~\cite{wu2026mcpmark}) probes adversarial behavior over 
generic toolsets. The closest antecedents at comparable scale are 
ITBench~\cite{jha2025itbench} (IT-operations) and 
AssetOpsBench~\cite{patel2025assetopsbenchbenchmarkingaiagents} 
(industrial asset operations), but neither uses MCP. Each of these 
makes evaluation choices that obscure the agent capabilities that 
matter for industrial deployment. \textbf{We identify three 
conflations:}
\begin{itemize}
    \item \textbf{The protocol conflation.} Non-MCP interfaces conflate
    \emph{protocol fluency} with \emph{reasoning ability}. An agent
    that fails could be failing at JSON-schema interpretation, not at
    PHM reasoning. PHMForge serves all tools through MCP, the protocol
    production agents will encounter.
    \item \textbf{The instrumentation conflation.} Synthetic-stub tools
    conflate \emph{agent failures} with \emph{instrumentation failures}:
    when a tool returns the wrong output, was the agent at fault or
    the tool? PHMForge's tools wrap published PHM algorithms
    (C-MAPSS-aligned RUL 
    estimators~\cite{saxena2008damage}, ISO~10816 vibration 
    analyzers~\cite{iso10816}),
    so failures attribute cleanly to reasoning rather than
    instrumentation.
    \item \textbf{The retrieval conflation.} Pre-specified tool sets
    conflate \emph{tool use} with \emph{tool retrieval}. Real
    industrial queries do not name the tools that should solve them.
    PHMForge's \emph{Unknown-Tools} mode evaluates retrieval as a
    first-class capability separate from invocation.
\end{itemize}

\paragraph{PHMForge as a methodological probe.}
We introduce \textbf{PHMForge}, a scenario-driven benchmark designed
as a methodological probe for industrial agentic AI
(Figure~\ref{fig:pipeline}). PHMForge stress-tests LLM agents on 
tool-grounded PHM reasoning through three domain-specific MCP 
servers exposing 39 algorithm-grounded tools. Agents must select, 
sequence, and compose diagnostic and maintenance tools, including 
in the Unknown-Tools mode. Because the MCP-native deployment 
surface is emerging in platforms such as 
Maximo~\cite{maximo_mcp_marketplace}, agents validated on PHMForge 
compose with production tooling without architectural changes. 
PHMForge is a reproducible proxy whose evaluation choices surface 
failure modes invisible to existing protocols, not a substitute 
for live deployment.

% --- Comparison table moved to end of Introduction (MCPMark layout) ---
\begin{table*}[t]
\centering
\small
\caption{PHMForge in context with eight peer benchmarks. \textit{PHM-Domain}. Tools and scenarios are physically grounded. \textit{Multi-Asset}. Covers $\geq$3 asset classes. \textit{Agentic}. Evaluates multi-turn tool orchestration. \textit{MCP-Native}. Tools served through MCP. \textit{Tool-Retrieval}. Agent must locate tools, not just invoke them. \textit{SME-Authored}. Scenarios authored by domain experts. \textit{Real Alg.} Tools wrap published algorithms rather than stubbed mocks. \textit{Det.\ Eval}. Deterministic.\protect\footnotemark[1]}
\label{tab:comparison}
\setlength{\tabcolsep}{4pt}
\renewcommand{\arraystretch}{1.15}
\begin{tabular}{l c c c c c c c c c}
\toprule
\textbf{Benchmark} &
\textbf{PHM-} & \textbf{Multi-} & \multirow{2}{*}{\textbf{Agentic}} &
\textbf{MCP-} & \textbf{Tool-} & \textbf{SME-} & \textbf{Real} & \textbf{Det.} &
\multirow{2}{*}{\textbf{\#Scen.}} \\
& \textbf{Domain} & \textbf{Asset} & & \textbf{Native} & \textbf{Retrieval} & \textbf{Authored} & \textbf{Alg.} & \textbf{Eval} & \\
\midrule
ITFormer~\cite{wang2025itformer}                                     & $\checkmark$ & $\times$     & $\times$     & $\times$     & $\times$     & $\times$     & $\times$     & $\checkmark$ & 110k \\
PDMBench~\cite{zhang2026pdmbench}                                   & $\checkmark$ & $\checkmark$ & $\times$     & $\times$     & $\times$     & $\times$     & $\times$     & $\checkmark$ & --   \\
PHM-Bench~\cite{yang2025phmbenchdomainspecificbenchmarkingframework} & $\checkmark$ & $\times$     & $\times$     & $\times$     & $\times$     & $\times$     & $\times$     & $\checkmark$ & --   \\
MLE-Bench~\cite{chan2025mlebenchevaluatingmachinelearning}           & $\times$     & --           & $\checkmark$ & $\times$     & $\times$     & $\times$     & $\times$     & $\checkmark$ & 75   \\
MCP-Bench~\cite{wang2025mcpbenchbenchmarkingtoolusingllm}            & $\times$     & --           & $\checkmark$ & $\checkmark$ & $\times$     & $\times$     & $\times$     & $\checkmark$ & 250  \\
MCP-Universe~\cite{luo2025mcpuniversebenchmarkinglargelanguage}      & $\times$     & --           & $\checkmark$ & $\checkmark$ & $\times$     & $\times$     & $\times$     & $\checkmark$ & 231  \\
ITBench~\cite{jha2025itbench}                                        & $\times$     & --           & $\checkmark$ & $\times$     & $\times$     & $\checkmark$ & $\times$     & $\checkmark$ & 121  \\
AssetOpsBench~\cite{patel2025assetopsbenchbenchmarkingaiagents}      & $\checkmark$ & Partial\protect\footnotemark[2]     & $\checkmark$ & $\times$     & $\times$     & $\checkmark$ & $\times$     & $\checkmark$ & 141  \\
\addlinespace[2pt]
\rowcolor{highlight}
\textbf{PHMForge} & $\checkmark$ & $\checkmark$ & $\checkmark$ & $\checkmark$ & $\checkmark$ & $\checkmark$ & $\checkmark$ & $\checkmark$ & \textbf{99} \\
\bottomrule
\end{tabular}
\vspace{-0.2in}
\end{table*}
\footnotetext[1]{PHMForge's deterministic evaluator applies to the automated ReAct/ReActXen harness on the 25-scenario stratified subset; the frontier evaluation (Claude Code, 99 scenarios) was conducted manually under the same scenario-level scoring rubric.}
\footnotetext[2]{AssetOpsBench covers anomaly detection and historical work-order analysis on a single asset class; PHMForge covers eight asset classes across rotating equipment, aero-engines, and lithium-ion battery storage.}

Table~\ref{tab:comparison} positions PHMForge against three research
threads. PHM benchmarks are domain-grounded but not agentic. Agent
benchmarks are agentic but not domain-grounded. MCP-native benchmarks
are protocol-aligned but operate over generic tools. The closest
peers at comparable scale are ITBench~\cite{jha2025itbench} (121
scenarios, IT-operations agents over non-MCP tools) and
AssetOpsBench~\cite{patel2025assetopsbenchbenchmarkingaiagents}
(141 scenarios, PHM domain with a synthetic
toolchain).\footnote{AssetOpsBench centers anomaly detection and
historical work-order analysis. PHMForge centers MCP-native tool
orchestration over the full PHM task taxonomy.} \emph{PHMForge is the
first benchmark to satisfy all seven design axes simultaneously},
which makes it a methodological probe in addition to a dataset.

\paragraph{Contributions.} 
Our contributions are: \emph{(i)} an MCP-native, algorithm-grounded 
PHM benchmark with 99 SME-authored scenarios across 8 asset 
classes and 5 task categories, served through 39 MCP tools across 
three domain-specific servers (no LLM in authoring; ground truth 
links to source citations; Krippendorff's $\alpha \in [0.74,\,0.82]$ 
on a 30-scenario stratified sample of the rotating-equipment and 
aero-engine subset); \emph{(ii)} data-grounded retrieval as a 
first-class evaluation axis via the Unknown-Tools mode (agents 
lose 21.3 pp pass@1 when they must autonomously identify and load 
the relevant dataset rather than receive it inline); \emph{(iii)} 
consistency-aware evaluation via the \emph{pass-all-3} metric 
(fraction solved on all three independent runs), the canonical 
measure for the lithium-ion battery architectural ablation 
(\S\ref{sec:ablations}); \emph{(iv)} a three-way failure taxonomy 
(reasoning, tool-invocation, orchestration) computed directly 
from execution traces; and \emph{(v)} reproducibility at academic 
cost (deterministic evaluators, public leaderboard, datasheet, 
open license, \$20--\$50 per run).

% =====================================================================
% SECTION 2 : PHMFORGE
% Combined section in MCPMark style: tool catalog → scenario
% construction → evaluation framework. Includes representative-scenarios
% figure and the sunburst+tools figure.
% =====================================================================
\section{PHMForge: An MCP-Native Industrial PHM Benchmark}
\label{sec:phmforge}

PHMForge has three components. The first is an MCP tool catalog 
grounded in published PHM algorithms (\S\ref{sec:tools}). The second 
is the set of 99 SME-authored scenarios across 8 asset classes 
spanning rotating equipment, aero-engines, and lithium-ion battery 
storage (\S\ref{sec:scenarios}). The third is an evaluation framework 
with two interaction modes, execution-based scoring, and 
trajectory-level diagnostics (\S\ref{sec:eval-framework}).

% --- Sunburst + tool list figure ---
\begin{figure*}[h!]
    \centering
    \includegraphics[width=1.05\textwidth]{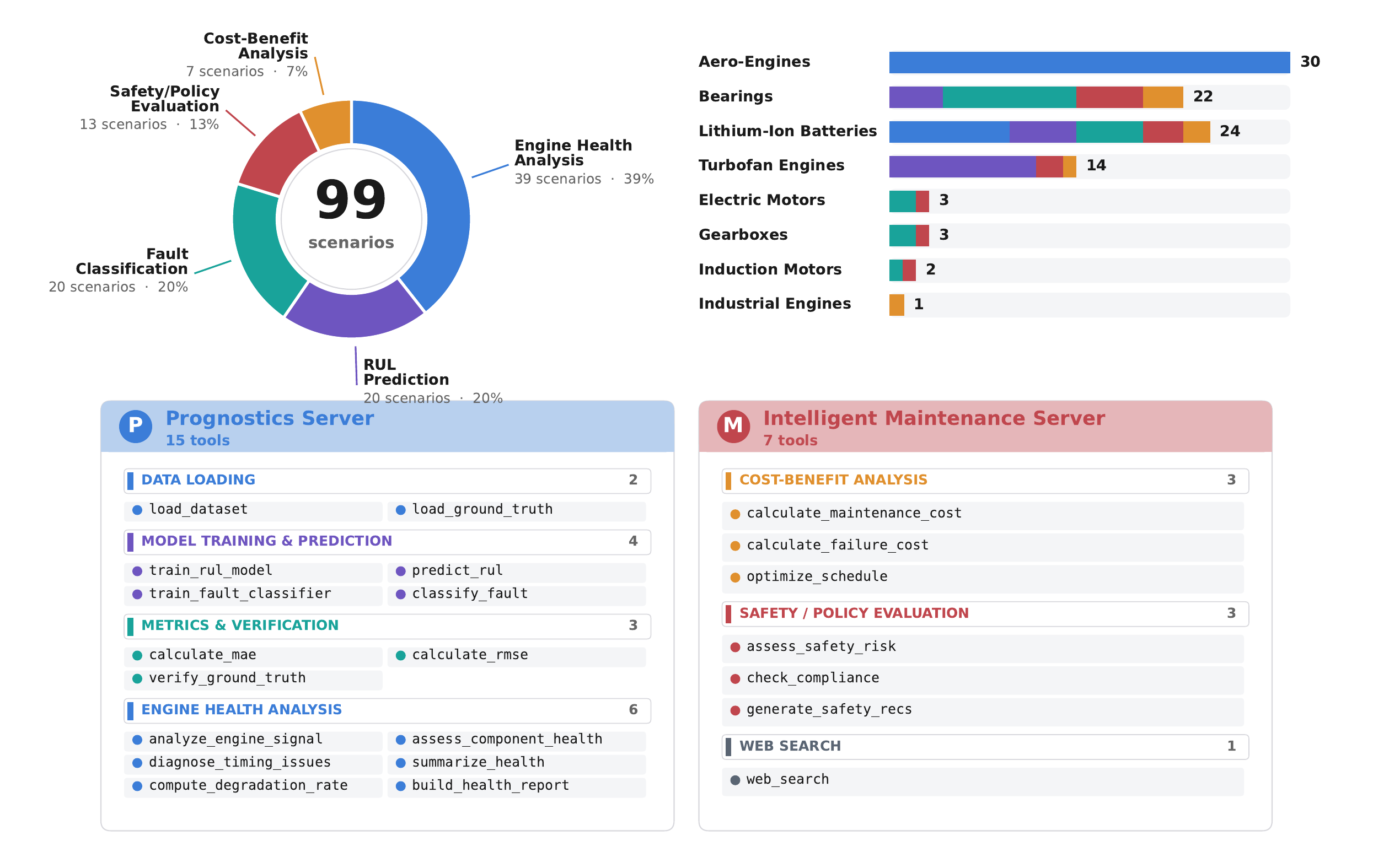}
    \vspace{-0.1in}
    \caption{\textbf{PHMForge benchmark composition.} 99 scenarios 
    across 5 task categories and 8 asset classes (top); 
    representative tools from the Prognostics and Maintenance servers 
    (bottom).}
    \label{fig:scenario_distribution}
    \vspace{-0.15in}
\end{figure*}
\subsection{Tool Catalog}
\label{sec:tools}

The public MCP ecosystem now covers digital workflows but exposes
none of the published PHM algorithms required to evaluate
domain-grounded reasoning. PHMForge therefore implements two
domain-specific servers (Figure~\ref{fig:scenario_distribution},
bottom). \emph{Every tool wraps an established PHM algorithm rather
than a synthetic stub}, so when an agent fails, the failure
attributes cleanly to reasoning rather than instrumentation. The
\textbf{Prognostics Server} implements C-MAPSS-aligned RUL
formulations~\cite{saxena2008damage} and ISO~10816 vibration
thresholds~\cite{iso10816}, with aero-engine component-level health
assessment over the Fan, LPC, HPC, HPT, and LPT modules. The
\textbf{Intelligent Maintenance Server} implements
preventive-versus-reactive cost decomposition, RUL-threshold
schedule optimization, and regulatory compliance against IEC~61508,
ISO~13849, OSHA, FAA, and NEMA. Required-toolset sizes per scenario
range from $|\mathcal{T}_\tau|=3$ to $7$ (mean $4.99$). Full
specifications appear in Appendix~\ref{app:tools}.

\paragraph{Battery Prognostics Server (17 tools).}
The lithium-ion BESS asset class was added in Stage~6 of the 
curation timeline (Appendix~\ref{app:curation-process}) and is 
served by a third domain-specific server providing 17 tools 
across four categories: data access (cycle, summary, impedance 
retrieval), diagnostics (capacity-SOH, z-score anomaly, thermal 
anomaly, impedance trend), prediction (linear regression, 
Arrhenius-aware empirical capacity fade, leave-one-battery-out 
LSTM, Chronos foundation model, TTM zero-shot and fine-tuned), 
and reporting (fleet-baseline comparison, end-to-end health 
reporting). TTM variants are exposed as separate tools to 
prevent silent mixing of training conditions; LSTM checkpoints 
are SHA256-fingerprinted to detect stale-cache reuse. Full 
specifications appear in Appendix~\ref{app:tools}.

\subsection{Scenario Construction}
\label{sec:scenarios}

\begin{figure*}[t]
    \centering
    \includegraphics[width=\textwidth]{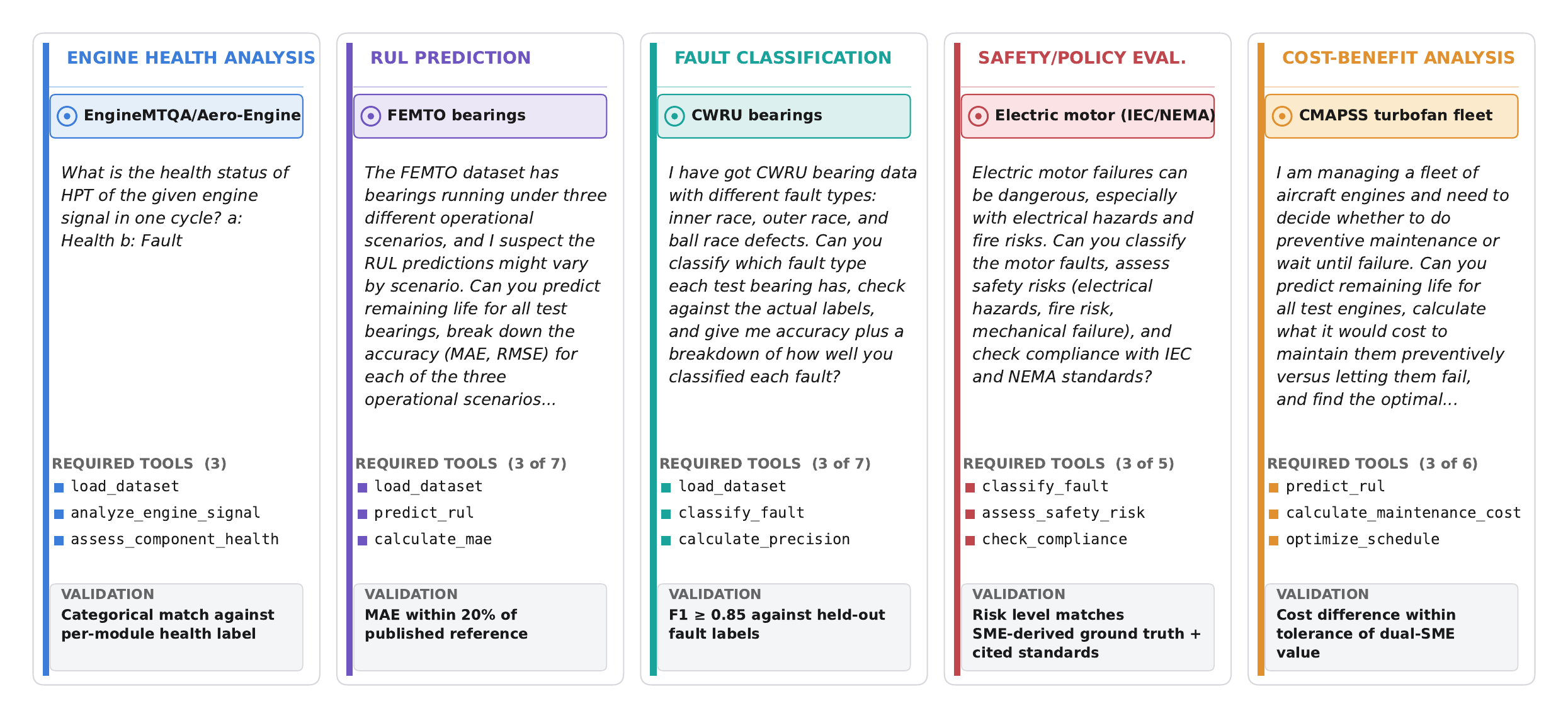}
    \caption{\textbf{Representative PHMForge scenarios across the
    five task categories.} Each card pairs an SME-authored
    natural-language query with a representative subset of the
    required tool set (full counts indicated, e.g., ``3 of 7'') and
    the validation criterion. Queries embed domain terminology
    (HPT, RUL, IEC, NEMA) as plant managers, technicians, and
    safety officers actually use it. Each scenario also includes
    2--4 distractor tools $\mathcal{T}_\tau^{-}$, omitted from the
    cards for clarity.}
    \label{fig:representative_scenarios}
\end{figure*}

Scenario design must balance three pressures: \emph{domain 
authenticity} (queries must reflect how operators phrase requests), 
\emph{evaluator determinism} (outputs must be machine-verifiable), 
and \emph{reproducibility under SME-labor constraints}. PHMForge 
resolves these through the protocol below; 
Figure~\ref{fig:representative_scenarios} shows five real 
scenarios, one per task category.

% --- Representative scenarios figure ---

\paragraph{Dataset and asset selection.}
We searched five public dataset 
platforms~\cite{kaggle_datasets,huggingface_datasets,UCI_ML_Repo,NASA_Prognostics_Data,PHM_Society_Archives} 
and retained 19 datasets across 8 asset classes after a 
three-stage filter (community validation, technical-quality, 
PHM-task alignment; Appendix~\ref{app:dataset-details}). Coverage 
spans aero-engines (C-MAPSS, EngineMT-QA), bearings (CWRU, HUST, 
FEMTO), electric/induction motors, gearboxes, industrial 
engines, turbofans, and lithium-ion battery cells 
(NASA PCoE B0005--B0018), and is intentionally biased toward 
asset domains with mature public datasets. Pumps, hydraulic 
systems, wind-turbine drivetrains, and HVAC systems remain 
unrepresented and are targets for community-driven expansion.

\paragraph{SME authoring.}
Following methodology adapted from
TabArena~\cite{erickson2025tabarenalivingbenchmarkmachine} and
MLE-Bench~\cite{chan2025mlebenchevaluatingmachinelearning},
scenarios were authored by a small SME consortium spanning
industrial asset specialists, a data scientist, and a
maintenance technician, with combined operational experience
across aerospace and rotating-equipment domains. Role
descriptions and the authoring workflow appear in
Appendix~\ref{app:curation-process}. \textbf{No LLM was used at any
point in scenario generation, query formulation, or ground-truth
derivation.} Authoring followed three guidelines:
\textit{(i)} queries incorporate domain-specific terminology and
acronyms as they appear in practice (HPC for High-Pressure
Compressor). \textit{(ii)} questions are framed in stakeholder
voices reflecting how plant managers, technicians, or safety
officers actually pose requests. \textit{(iii)} tool names and task
intentions deliberately avoid lexical overlap, preventing
surface-level keyword matching. Figure~\ref{fig:representative_scenarios}
shows the result. Queries embed terms like \emph{HPT health status},
\emph{three operational scenarios}, and \emph{IEC and NEMA
standards} naturally, without naming the tools that resolve them.
\textbf{Every scenario was end-to-end run by a human SME using the
tool catalog before release}. Scenarios that could not be solved
with the published tools were either revised or rejected.

\paragraph{Scenario structure.}
Every scenario is a tuple
$\tau = (\mathcal{Q},\,\mathcal{D},\,\mathcal{T}_\tau,\,
\mathcal{T}_\tau^{-},\,\mathcal{G})$:
the natural-language query~$\mathcal{Q}$, dataset
context~$\mathcal{D}$, required tool subset~$\mathcal{T}_\tau$,
distractor tool set~$\mathcal{T}_\tau^{-}$ (plausible but
task-inappropriate tools), and task-specific ground
truth~$\mathcal{G}$. Output templates are task-specific. Continuous
numerical fields for RUL Prediction, discrete categorical strings
for Fault Classification, and multiple-choice or categorical labels
for Engine Health Analysis (see the EngineMTQA card in
Figure~\ref{fig:representative_scenarios}).

\paragraph{Inter-annotator agreement.}
We computed inter-annotator agreement (IAA) on a stratified
30-scenario sample drawn from the 75 rotating-equipment and
aero-engine scenarios. The 24 lithium-ion battery scenarios
constituting the BESS extension were authored, dual-reviewed, and
SME-executed under the same procedural protocol, but
Krippendorff's $\alpha$ is not reported on this subset; the
reported IAA values therefore cover the rotating-equipment and
aero-engine scope, not the full 99-scenario benchmark. We flag
this as a known scope limitation
(\S\ref{sec:limitations} and
Appendix~\ref{app:iaa-bess}). On the rated 30-scenario sample,
two SMEs scored each scenario across three dimensions on a
4-point Likert scale. Per-dimension Krippendorff's $\alpha$
exceeds the conventional $\alpha=0.7$ threshold for substantial
agreement on every rated dimension
(Table~\ref{tab:iaa})~\cite{krippendorff2018content}. Of the 30
dual-rated scenarios, 7 had a disagreement of 2 or more Likert
points on at least one dimension. These entered a structured
resolution protocol with a third consortium member and were either
revised or replaced. Confidence intervals, the LLM-as-judge
cross-check, and the resolution protocol appear in
Appendix~\ref{app:annotation}.

\begin{table}[t]
\caption{Inter-annotator agreement (Krippendorff's $\alpha$) on
the 30-scenario stratified sample drawn from the 75
rotating-equipment and aero-engine scenarios. The 24 lithium-ion
battery scenarios are not covered by this IAA study (see
Appendix~\ref{app:iaa-bess}). CIs are 95\% bootstrap intervals
(1{,}000 resamples, full details in Appendix~\ref{app:annotation}).}
\label{tab:iaa}
\centering
\small
\begin{tabular}{lcc}
\toprule
\textbf{Dimension}        & \textbf{Krippendorff's $\alpha$} & \textbf{95\% CI} \\
\midrule
Realism                   & 0.78 & [0.71, 0.85] \\
Difficulty calibration    & 0.74 & [0.66, 0.81] \\
Ground-truth correctness  & 0.82 & [0.75, 0.88] \\
Pooled                    & 0.78 & [0.73, 0.83] \\
\bottomrule
\end{tabular}
\end{table}

\paragraph{Ground-truth construction.}
Where the source dataset provides ground truth directly (RUL
trajectories in C-MAPSS, labeled fault taxonomies in CWRU), we
extract programmatically and validate against published benchmark
numbers. For threshold or compliance judgments (Cost-Benefit,
Safety/Policy), SMEs derive ground truth from cited literature and
standards documents, with a second SME independently re-deriving
the same value before the scenario enters the benchmark. Only
concordant dual derivations are retained. \textbf{Every
\texttt{ground\_truth} field links to a source citation or
extraction script, making the entire ground-truth set auditable.}
Outputs are scored with task-commensurate metrics. MAE/RMSE for
RUL Prediction, accuracy/precision/recall/F1 for Fault
Classification, and categorical matching for Engine Health
Analysis. The methodology has clearer empirical footing for tasks
with directly verifiable ground truth (RUL Prediction and Fault
Classification: 60 of 99 scenarios) than for those requiring
threshold judgments (Cost-Benefit and Safety/Policy: 20 of 99). We
discuss this scoping limitation in \S\ref{sec:limitations}.

\paragraph{Threats to construct validity.}
Because the SME consortium authored both the tool wrappers and 
the scenarios, scenarios that exposed gaps in the tool set were 
revised or rejected (\S\ref{sec:scenarios}). This biases PHMForge 
toward measuring orchestration over an existing tool surface 
rather than identifying when no tool suffices. The Unknown-Tools 
mode (\S\ref{sec:eval-framework}) and cross-equipment transfer 
protocol (\S\ref{sec:ablations}) partially mitigate this; residual 
circularity is best read as a ceiling. A related Goodhart-style 
risk is that ground truth for RUL Prediction and Fault 
Classification is derived from the same algorithms exposed as 
MCP tools, flagged in \S\ref{sec:limitations}.

\subsection{Evaluation Framework}
\label{sec:eval-framework}

\paragraph{Interaction modes and scoring.}
PHMForge supports two modes. \textbf{Tools-Provided}: the agent 
receives $\mathcal{T}_\tau \cup \mathcal{T}_\tau^{-}$ and must select, 
sequence, and invoke the correct subset, isolating tool-orchestration 
capability. \textbf{Unknown-Tools}: the agent receives only 
$\mathcal{Q}$ and must retrieve tools from the full server catalog 
\emph{and} identify the relevant dataset before invoking them, 
isolating data-discovery, tool-retrieval, and intent-extraction; 
the data-discovery component is reported quantitatively in 
Appendix~\ref{app:additional-ablations}. 
An agent $\mathcal{A}$ paired with model $\mathcal{M}$ produces 
output $\hat{y}$ and trajectory 
$\pi = \langle (t_1, a_1), \ldots, (t_k, a_k) \rangle$; success is 
binary, $E(\mathcal{M}, \mathcal{A}, \tau) = \mathbb{1}[\text{validate}(\hat{y}, \mathcal{G})]$. 
Each scenario runs three times at $T=0$. We report \textbf{pass@1} 
(mean success, capability) and \textbf{pass-all-3} (fraction solved 
on every run, consistency, analogous to MCPMark's 
pass\textsuperscript{4}~\cite{wu2026mcpmark}); an agent that 
succeeds sometimes but not always cannot be deployed in 
safety-critical settings. With $T=0$ decoding, residual 
across-run variance reflects API non-determinism rather than 
stochastic sampling, and is interpreted as a robustness lower 
bound.

\paragraph{Failure decomposition and reproducibility.}
We complement binary $E(\cdot)$ with three trajectory-level 
categories: \textbf{reasoning errors} (incoherent plans, 
distractor invocations, task misclassification), 
\textbf{tool-invocation errors} (malformed arguments, type 
mismatches, schema failures), and \textbf{orchestration errors} 
(correct calls assembled with wrong sequencing, state-dependency 
violations, premature termination), plus tool-precision/recall 
over $\mathcal{T}_\tau$, sequencing accuracy, and trajectory 
length. Categorization is computed deterministically from MCP 
execution logs, making failure decompositions reproducible from 
the released traces. Every run produces a deterministic JSON 
record in a sandboxed container with pinned dependencies. A complete benchmark run 
costs \$20--\$50 per (framework, model) combination, two orders 
of magnitude below comparable agentic 
benchmarks~\cite{chan2025mlebenchevaluatingmachinelearning,luo2025mcpuniversebenchmarkinglargelanguage}. 
We follow the Datasheets-for-Datasets 
standard~\cite{gebru2021datasheets}.

% =====================================================================
% SECTION 3 : EXPERIMENTS AND RESULTS
% Cross-framework with pass-all-3 added → process metrics → task
% specific → failure pies → ablations.
% =====================================================================

\section{Experiments and Results}
\label{sec:results}

We organize the experimental study around four questions a 
practitioner cares about. First, among open-source agentic 
frameworks paired with open-weight LLMs, which combination performs 
best, and is the result stable across runs (\S\ref{sec:cross-framework})? 
Second, when we lift the budget constraint and use frontier models 
across the full benchmark, how close are agents to a 
production-deployment threshold (\S\ref{sec:frontier-99})? Third, 
can a state-of-the-art LLM substitute for human SME consensus when 
scoring PHMForge scenarios (\S\ref{sec:llm-judge})? Fourth, where 
do failures concentrate, and would alternative architectures 
(text-RAG, free-form code, cross-equipment transfer) close the gap 
(\S\ref{sec:ablations}--\S\ref{sec:task-metrics})? 

\subsection{Framework Comparison: ReAct vs ReActXen}
\label{sec:cross-framework}

\paragraph{Setup.}
A PHMForge \emph{agent} pairs an agentic framework (ReAct, 
ReActXen, or Claude Code) with an LLM backbone that orchestrates 
MCP tool calls. To stay within compute budget, we draw a 
25-scenario stratified subset preserving all five task categories 
and run six API-served open-weight backbones: Llama-3.3-70B, 
Llama-4-Maverick-17B-128E MoE, Mistral-Medium-2505, 
Mistral-Small-3.1-24B, GPT-OSS-120B, and a compact open-weight 
LLM~\cite{granite}.

\paragraph{Findings.}
The strongest configuration is \textbf{ReAct + 
Llama-4-Maverick-17B-128E} at \textbf{80.0\% Pass@1}, which we 
carry forward into per-task and ablation analyses 
(\S\ref{sec:ablations}--\S\ref{sec:task-metrics}). Three 
patterns emerge from Table~\ref{tab:framework_performance}. 
\textit{(i)~Single-pass ReAct outperforms iterative 
ReAct+Reflection on most backbones} (80.0\% vs.\ 63.6\% on 
Llama-4-Maverick): on well-bounded tool-orchestration tasks 
extra reflection more often hallucinates tool calls than 
corrects them; the exception is GPT-OSS-120B (56\%$\to$68\%) 
where reflection corrects initial tool-routing errors. 
\textit{(ii)~Long-context tool-input formatting bottlenecks 
RUL.} Mistral-Medium-2505 truncates 100-element arrays passed 
to \texttt{calculate\_mae} (0\% RUL), while Llama-4-Maverick 
chunks and externally aggregates (60\%). \textit{(iii)~Mid-tier 
models excel at structured judgment but underperform at numerical 
orchestration:} Mistral-Medium-2505 reaches 100\% on Cost-Benefit 
and Safety/Policy yet 0\% on RUL, suggesting 
that benchmarking only on discriminative tasks would substantially 
overstate predictive maintenance capability.

\begin{table*}[t]
\caption{Framework-and-model Pass@1 across PHM task categories on a 
25-scenario stratified subset (5 RUL + 5 Fault + 10 Health + 2 Cost 
+ 3 Safety) preserving all 5 task categories. $^\dagger$Partial 
coverage on Cost-Benefit and Safety/Policy categories; 
ReActXen's reflection loop on Llama-4-Maverick exceeded the 
compute budget on those scenarios. Best per column in \textbf{bold}.}
\label{tab:framework_performance}
\centering
\small
\setlength{\tabcolsep}{4pt}
\begin{tabular}{@{}lcccccc@{}}
\toprule
\textbf{Framework + Model} & \textbf{Pass@1} & \textbf{RUL} & \textbf{Fault} & \textbf{Health} & \textbf{Cost} & \textbf{Safety} \\
& & (5) & (5) & (10) & (2) & (3) \\
\midrule
\rowcolor{highlight}
\textbf{ReAct + Llama 4 Maverick} & \textbf{80.0\%} & 60.0\% & \textbf{100.0\%} & 80.0\% & 50.0\% & \textbf{100.0\%} \\
ReAct + Mistral Medium 2505 & 64.0\% & 0.0\% & 40.0\% & \textbf{90.0\%} & \textbf{100.0\%} & \textbf{100.0\%} \\
ReAct + GPT-OSS 120B & 56.0\% & 0.0\% & \textbf{100.0\%} & 50.0\% & 50.0\% & \textbf{100.0\%} \\
ReAct + Compact-LLM & 44.0\% & 20.0\% & 40.0\% & 60.0\% & 50.0\% & 33.3\% \\
ReAct + Mistral Small 3.1 24B & 44.0\% & 40.0\% & 60.0\% & 30.0\% & 50.0\% & 66.7\% \\
ReAct + Llama 3.3 70B & 36.0\% & 20.0\% & 20.0\% & 20.0\% & \textbf{100.0\%} & \textbf{100.0\%} \\
\midrule
ReActXen + GPT-OSS 120B & 68.0\% & 20.0\% & \textbf{100.0\%} & \textbf{90.0\%} & 50.0\% & 33.3\% \\
ReActXen + Llama 4 Maverick$^\dagger$ & 63.6\% & 20.0\% & \textbf{100.0\%} & \textbf{100.0\%} &  &  \\
ReActXen + Compact-LLM & 48.0\% & 20.0\% & 0.0\% & 60.0\% & \textbf{100.0\%} & \textbf{100.0\%} \\
ReActXen + Mistral Medium 2505 & 48.0\% & 20.0\% & 40.0\% & 50.0\% & \textbf{100.0\%} & 66.7\% \\
ReActXen + Mistral Small 3.1 24B & 48.0\% & 40.0\% & 40.0\% & 30.0\% & \textbf{100.0\%} & \textbf{100.0\%} \\
\bottomrule
\end{tabular}
\end{table*}

\subsection{Frontier Model Performance on the Full 99-Task Benchmark}
\label{sec:frontier-99}

We then lift the cost-budget constraint and evaluate Claude Code 
paired with Claude Sonnet~4.5 and Claude Opus~4.6 on \emph{all} 99 
PHMForge scenarios. Claude Code is an interactive CLI agent, not 
API-callable from the automated harness, so this configuration 
is evaluated manually under the same scenario-level scoring. The 
intent is to characterize how close current frontier agents come 
to a deployment-grade threshold; industrial adoption studies 
typically cite $\sim$85\% task accuracy as a precondition for 
unsupervised operation. Opus~4.6 reaches \textbf{80.8\% pass@1} 
across the 99 scenarios and Sonnet~4.5 reaches 64.6\% 
(Table~\ref{tab:frontier_performance}). These figures are not 
directly comparable to the \S\ref{sec:cross-framework} 
open-weight numbers (different harness, n=25 stratified subset); 
we treat the two regimes as separate experiments. Both leave a 
4.2--20.4 point gap below 85\%, motivating the failure analysis 
that follows.

\begin{table*}[t]
\caption{Frontier-model Pass@1 across PHM task categories on the 
full 99-scenario benchmark (39 Health + 20 RUL + 20 Fault + 13 
Safety + 7 Cost). Manual evaluation under scenario-level scoring; 
Claude Code is the interactive CLI agent paired with each frontier 
backbone. Best per column in \textbf{bold}.}
\label{tab:frontier_performance}
\centering
\small
\setlength{\tabcolsep}{4pt}
\begin{tabular}{@{}lcccccc@{}}
\toprule
\textbf{Framework + Model} & \textbf{Pass@1} & \textbf{RUL} & \textbf{Fault} & \textbf{Health} & \textbf{Cost} & \textbf{Safety} \\
& & (20) & (20) & (39) & (7) & (13) \\
\midrule
\rowcolor{highlight}
\textbf{Claude Code + Opus 4.6}    & \textbf{80.8\%} & \textbf{85.0\%} & \textbf{75.0\%} & \textbf{79.5\%} & \textbf{71.4\%} & \textbf{92.3\%} \\
Claude Code + Sonnet 4.5            & 64.6\% & 70.0\% & 60.0\% & 69.2\% & 42.9\% & 61.5\% \\
\bottomrule
\end{tabular}
\end{table*}

\subsection{LLM-as-Judge vs.\ Human Evaluation}
\label{sec:llm-judge}

A state-of-the-art LLM judge (Claude 
Sonnet~4.0~\cite{claude_sonnet_docs}) over the same 30-scenario 
sample with identical rubrics yields Krippendorff's 
$\alpha = 0.61$, well below human--human agreement 
($\alpha \in [0.74,\,0.82]$); the LLM over-rated realism and 
under-rated difficulty calibration 
(Appendix~\ref{app:annotation}). We therefore use human SME 
consensus as PHMForge's canonical scoring authority and treat 
LLM-as-judge as unreliable for difficulty assessment in domains 
requiring deep expertise.

\subsection{Architectural Ablations}
\label{sec:ablations}

\paragraph{MCP vs.\ text-based RAG.}
On 24 lithium-ion battery scenarios under three independent runs 
at $T=0$ with Claude Opus~4.6, replacing MCP execution (Path~B) 
with a Chroma-indexed RAG pipeline (Path~A) drops mean pass@1 
from 80.6\% to 48.6\% on operator-style queries and from 91.7\% 
to 73.6\% on protocol-style queries (Wilson 95\% CIs and McNemar 
tests in Table~\ref{tab:bess-stability}). The operator-style gap 
is highly significant ($p{=}0.002$), and Path~A never out-passes 
Path~B under loose phrasing on this set; the protocol-style gap 
is marginal ($p{=}0.07$). On RUL Prediction the collapse is 
sharpest: Path~A drops to 1/5 pass-all-3 while Path~B retains 
5/5, a \textbf{100\%~$\to$~20\% pass-all-3 collapse}. 
Per-category breakdowns appear in 
Appendix~\ref{app:bess-pass-all-3}.

\paragraph{With vs.\ without domain tools.}
With Claude Code + Opus~4.6, withholding all domain-specific MCP 
tools and forcing the agent to rely on native execution drops 
aggregate completion from 80.8\% to 25\%, a \textbf{56-point 
collapse} confirming PHMForge measures orchestration over 
algorithm-grounded tools rather than open-ended coding. Tool 
subsets and per-task counts are in the supplementary ablation 
summary.

\paragraph{Cross-equipment transfer.}
In-distribution scenarios reach 84.1\% completion; zero-shot 
transfer from bearing diagnostics to motor diagnostics collapses 
to 42.7\%, a \textbf{41-point gap} despite shared PHM task 
structure. Additional ablations (ground-truth verification, 
distractor tools, data-discovery under Unknown-Tools mode) appear 
in Appendix~\ref{app:additional-ablations}.

\subsection{Failure Decomposition}
\label{sec:failure-decomp}

Three findings stand out from projecting per-task failures onto 
the three-way taxonomy of \S\ref{sec:eval-framework}. (i) 
\textbf{Orchestration errors dominate}: most failures involve 
correct individual tool calls in the wrong order, consistent 
with a 23\% trajectory-level incorrect-sequencing rate; frontier 
LLMs are stronger at \emph{calling} tools than at \emph{planning 
when to call them}. (ii) \textbf{Tool-invocation errors decline 
with backbone capacity}: schema-invalid calls are concentrated 
in smaller open-weight models and become rare in frontier 
configurations, suggesting schema-validated MCP shifts error 
modes upward as capacity grows. (iii) On the lithium-ion battery 
subset, \textbf{Cost-Benefit failures (0/2 across both Opus and 
Sonnet) reflect ambiguous fuzzy queries lacking budget anchors} 
rather than orchestration errors. Trajectory-level metrics are 
in Appendix~\ref{app:process-metrics}.

\subsection{Task-Specific Performance and Quality Metrics}
\label{sec:task-metrics}

Task-commensurate quality metrics for the strongest configuration 
(Claude Code + Opus~4.6) are reported in 
Table~\ref{tab:task_metrics}. The agent reaches deployment-grade 
performance on Safety/Policy and Cost-Benefit ($|\Delta|\le 4$\,pp 
from SME consensus) but lags published baselines on RUL Turbofan 
(MAE +4.3\,cy) and on Motor fault discrimination 
(--12.2\,pp accuracy). On the lithium-ion battery class, the TTM 
fine-tuned predictor reaches \textbf{13.5-cycle MAE} under 
leave-one-battery-out evaluation, outperforming linear regression 
(28.4\,cy) and Chronos fine-tuned (31.8\,cy) by $\sim$50\% under 
the same protocol (Appendix~\ref{app:bess-rul-benchmark}). 
Per-task error analysis (Appendix~\ref{app:gt-eval}) traces these 
gaps to train-test contamination, invalid range predictions, and 
cross-equipment generalization failures.

\begin{table*}[!ht]
\caption{\textbf{Task-specific quality metrics} for the strongest 
configuration (Claude Code + Opus~4.6). The agent matches reference 
baselines on Safety/Policy and Cost-Benefit ($|\Delta|\le 4$\,pp) 
but lags on Motor fault discrimination ($|\Delta|\ge 11$\,pp). 
$\downarrow$/$\uparrow$ indicates direction of improvement; baselines 
from PHM-Bench~\cite{yang2025phmbenchdomainspecificbenchmarkingframework} 
for RUL and Fault Classification or SME consensus for Decision tasks. 
$^*$Lithium-Ion RUL evaluated under leave-one-battery-out (LOO) 
protocol with the TTM fine-tuned predictor; baseline is 
linear regression under the same protocol (Appendix~\ref{app:bess-rul-benchmark}).}
\label{tab:task_metrics}
\centering
\small
\setlength{\tabcolsep}{6pt}
\renewcommand{\arraystretch}{1.05}
\begin{tabular}{@{}llrrlll@{}}
\toprule
\textbf{Asset} & \textbf{Metric} & \textbf{Agent} & \textbf{Baseline} & \textbf{$\Delta$} & \textbf{Primary Error Mode} & \textbf{Rate} \\
\midrule
\multicolumn{7}{@{}l}{\textit{\textbf{RUL Prediction}}} \\
\addlinespace[1pt]
\rowcolor{gray!6}
Turbofan & MAE $\downarrow$ (cy)  & 14.8 & \textcolor{gray!70}{10.5} & $+4.3\,\uparrow$ & Train-test contamination & 27\% \\
\rowcolor{gray!6}
         & RMSE $\downarrow$ (cy) & 19.2 & \textcolor{gray!70}{13.8} & $+5.4\,\uparrow$ & & \\
Bearing  & MAE $\downarrow$ (\%)  & 11.3 & \textcolor{gray!70}{8.2}  & $+3.1\,\uparrow$ & Invalid range predictions & 15\% \\
         & RMSE $\downarrow$ (\%) & 14.7 & \textcolor{gray!70}{10.9} & $+3.8\,\uparrow$ & & \\
\rowcolor{gray!6}
Lithium-Ion & MAE $\downarrow$ (cy)$^*$  & \textbf{13.5} & \textcolor{gray!70}{28.4} & $-14.9\,\downarrow$ & Capacity regeneration & --- \\
\addlinespace[2pt]
\midrule
\multicolumn{7}{@{}l}{\textit{\textbf{Fault Classification}}} \\
\addlinespace[1pt]
\rowcolor{gray!6}
Bearing & F1 $\uparrow$        & 0.84 & \textcolor{gray!70}{0.87} & $-0.03\,\downarrow$ & Fine-grained taxonomy & 23\% \\
\rowcolor{gray!6}
        & Acc $\uparrow$ (\%)  & 87.2 & \textcolor{gray!70}{91.3} & $-4.1\,\downarrow$  & & \\
Motor   & F1 $\uparrow$        & 0.68 & \textcolor{gray!70}{0.79} & $-0.11\,\downarrow$ & Severity distinction & 31\% \\
        & Acc $\uparrow$ (\%)  & 71.4 & \textcolor{gray!70}{83.6} & $-12.2\,\downarrow$ & & \\
\addlinespace[2pt]
\midrule
\multicolumn{7}{@{}l}{\textit{\textbf{Decision \& Compliance Tasks}}} \\
\addlinespace[1pt]
\rowcolor{gray!6}
\multicolumn{2}{@{}l}{Engine Health $\uparrow$ (\%)}  & 72.0 & \textcolor{gray!70}{78.6\,(SME)} & $-6.6\,\downarrow$ & Timing misdiagnosis  & 23\% \\
\multicolumn{2}{@{}l}{Safety / Policy $\uparrow$ (\%)}    & 90.0 & \textcolor{gray!70}{94.0\,(SME)} & $-4.0\,\downarrow$ & Threshold lookup     & 10\% \\
\rowcolor{gray!6}
\multicolumn{2}{@{}l}{Cost-Benefit (ROI window)} & $\pm 9.0\%$ & \textcolor{gray!70}{$\pm 5.0\%$\,(SME)} & $+4.0$\,pp wider & Cost-parameter misuse & 12\% \\
\bottomrule
\end{tabular}
\end{table*}

% =====================================================================
% SECTION 4 : LIMITATIONS
% =====================================================================
\vspace{-0.4em}
\section{Limitations and Future Work}
\label{sec:limitations}
\vspace{-0.3em}

PHMForge inherits the multi-call overhead intrinsic to the MCP 
paradigm~\cite{wu2026mcpmark}: repeated client--server round-trips 
amplify inference latency and token consumption, motivating future 
work in high-performance agent serving infrastructure.

% =====================================================================
% SECTION 5 : CONCLUSION
% =====================================================================
\vspace{-0.8em}
\section{Conclusion}
\label{sec:conclusion}
\vspace{-0.5em}

PHMForge is an MCP-native benchmark for industrial PHM where 
frontier LLMs reach 80.8\% pass@1 with the residual gap 
concentrated in orchestration errors. Replacing MCP tools with 
text-based RAG collapses RUL pass-all-3 from 100\% to 20\% on 
the lithium-ion battery class, and withholding domain tools 
entirely drops completion to 25\%; algorithm-grounded MCP tools 
are a necessary substrate for industrial deployment, not a 
saturated leaderboard.

\bibliographystyle{plainnat}
\bibliography{reference_kdd}

% ==============================================================
% APPENDIX (cleaned: removed duplicate sections, legacy results)
% Modeled on arxiv 2604.01532 structure: A. Tools, B. Datasets,
% C. Extended GT/Eval, with D. Curation, E. IAA, F. BESS added.
% ==============================================================
\appendix

\section{Tool Specifications}
\label{app:tools}

PHMForge exposes 39 specialized tools via three Model Context Protocol (MCP) servers. Tables~\ref{tab:prognostics_tools},~\ref{tab:maintenance_tools}, and~\ref{tab:battery_tools} provide complete specifications including parameter signatures for reproducibility.

\begin{table*}[!ht]
\caption{Complete tool inventory for the \textbf{Prognostics Server} (\texttt{prognostics-server}), providing 15 tools for data loading, model training, prediction, metric computation, and engine health analysis.}
\label{tab:prognostics_tools}
\centering
\normalsize
\resizebox{\textwidth}{!}{%
\begin{tabular}{@{}llp{6.5cm}@{}}
\toprule
\textbf{Tool Name} & \textbf{Description} & \textbf{Key Parameters} \\
\midrule
\multicolumn{3}{l}{\textbf{\textit{Data Loading}}} \\
\texttt{load\_dataset} & Load dataset from PDMBench data directory & \texttt{dataset\_name} (str), \texttt{split} (str, default=\texttt{"train"}) \\
\texttt{load\_ground\_truth} & Load ground truth RUL values or fault labels & \texttt{dataset\_name} (str), \texttt{file} (str, optional) \\
\midrule
\multicolumn{3}{l}{\textbf{\textit{Model Training}}} \\
\texttt{train\_rul\_model} & Train RUL prediction model with Adam optimizer & \texttt{dataset} (str), \texttt{model\_type} (str: mlp/lstm/transformer), \texttt{epochs} (int, default=50) \\
\texttt{train\_fault\_classifier} & Train multi-class fault classification model & \texttt{dataset} (str), \texttt{model\_type} (str), \texttt{epochs} (int) \\
\midrule
\multicolumn{3}{l}{\textbf{\textit{Prediction}}} \\
\texttt{predict\_rul} & Predict RUL for test units & \texttt{model\_path} (str), \texttt{test\_data} (str), \texttt{unit\_id} (int) \\
\texttt{classify\_faults} & Classify faults for test units & \texttt{model\_path} (str), \texttt{test\_data} (str), \texttt{unit\_id} (int) \\
\midrule
\multicolumn{3}{l}{\textbf{\textit{Metrics}}} \\
\texttt{calculate\_mae} & Calculate Mean Absolute Error for RUL predictions & \texttt{ground\_truth} (str), \texttt{predictions} (str) \\
\texttt{calculate\_rmse} & Calculate Root Mean Squared Error for RUL predictions & \texttt{ground\_truth} (str), \texttt{predictions} (str) \\
\texttt{verify\_ground\_truth} & Verify predictions against ground truth RUL values & \texttt{ground\_truth} (str), \texttt{predictions} (str) \\
\texttt{calculate\_accuracy} & Calculate classification accuracy for fault classification & \texttt{ground\_truth} (str), \texttt{predictions} (str) \\
\texttt{verify\_classification} & Verify fault classifications against ground truth & \texttt{ground\_truth} (str), \texttt{predictions} (str) \\
\midrule
\multicolumn{3}{l}{\textbf{\textit{Engine Health Analysis}}} \\
\texttt{analyze\_engine\_signals} & Parse multi-sensor signal data and identify anomalies & \texttt{sensor\_data} (str, JSON), \texttt{engine\_id} (str, optional) \\
\texttt{assess\_component\_health} & Evaluate health of turbofan components (Fan/LPC/HPC/HPT/LPT) & \texttt{component} (str), \texttt{efficiency} (float), \texttt{flow\_modifier} (float) \\
\texttt{diagnose\_timing\_issues} & Identify efficiency vs.\ flow-modifier degradation & \texttt{efficiency\_deviation} (float), \texttt{flow\_deviation} (float) \\
\texttt{detect\_degradation\_trend} & Detect degradation patterns over cycles & \texttt{cycle\_data} (str, JSON array) \\
\bottomrule
\end{tabular}%
}
\end{table*}

\begin{table*}[!ht]
\caption{Complete tool inventory for the \textbf{Maintenance Server} (\texttt{maintenance-server}), providing 7 tools for cost-benefit analysis, safety/policy evaluation, and web search.}
\label{tab:maintenance_tools}
\centering
\normalsize
\resizebox{\textwidth}{!}{%
\begin{tabular}{@{}llp{6.5cm}@{}}
\toprule
\textbf{Tool Name} & \textbf{Description} & \textbf{Key Parameters} \\
\midrule
\multicolumn{3}{l}{\textbf{\textit{Cost-Benefit Analysis}}} \\
\texttt{calculate\_maintenance\_cost} & Compute annual preventive maintenance costs including downtime & \texttt{preventive\_cost} (float), \texttt{frequency\_per\_year} (int), \texttt{downtime\_hours} (float), \texttt{hourly\_rate} (float) \\
\texttt{calculate\_failure\_cost} & Estimate expected annual cost of unplanned failures & \texttt{failure\_probability} (float), \texttt{repair\_cost} (float), \texttt{downtime\_hours} (float), \texttt{hourly\_rate} (float), \texttt{consequential\_cost} (float) \\
\texttt{optimize\_maintenance\_schedule} & Find cost-optimal RUL threshold for scheduling maintenance & \texttt{rul\_estimate} (float), \texttt{preventive\_cost} (float), \texttt{failure\_cost} (float), \texttt{safety\_margin} (float) \\
\midrule
\multicolumn{3}{l}{\textbf{\textit{Safety \& Policy Evaluation}}} \\
\texttt{assess\_safety\_risk} & Classify risk level (low/medium/high/critical) using RPN analysis & \texttt{failure\_mode} (str), \texttt{severity} (int, 1--10), \texttt{probability} (int, 1--10), \texttt{detectability} (int, 1--10) \\
\texttt{check\_compliance} & Validate against IEC/ISO/OSHA safety standards & \texttt{standard} (str), \texttt{safety\_integrity\_level} (int, 1--4), \texttt{current\_pfd} (float) \\
\texttt{generate\_safety\_recommendations} & Produce prioritized safety action items based on risk assessment & \texttt{risk\_level} (str), \texttt{failure\_mode} (str), \texttt{current\_controls} (str) \\
\midrule
\multicolumn{3}{l}{\textbf{\textit{Web Search}}} \\
\texttt{web\_search} & Search the internet for domain-specific information via Brave Search API & \texttt{query} (str), \texttt{count} (int) \\
\bottomrule
\end{tabular}%
}
\end{table*}

\begin{table*}[h!]
\caption{Complete tool inventory for the \textbf{Battery Prognostics Server} (\texttt{battery-prognostics-server}), providing 17 tools for the lithium-ion battery storage asset class. See Appendix~\ref{app:bess} for methodology details.}
\label{tab:battery_tools}
\centering
\normalsize
\resizebox{\textwidth}{!}{%
\begin{tabular}{@{}llp{6.5cm}@{}}
\toprule
\textbf{Tool Name} & \textbf{Description} & \textbf{Key Parameters} \\
\midrule
\multicolumn{3}{l}{\textbf{\textit{Data Access}}} \\
\texttt{fetch\_cycle\_data}    & Full per-cycle telemetry with all sensor channels & \texttt{battery\_id} (str), \texttt{cycle} (int) \\
\texttt{fetch\_cycle\_summary} & 5-feature lightweight projection (capacity, temp, voltage, duration) & \texttt{battery\_id} (str), \texttt{cycle\_range} (tuple) \\
\texttt{fetch\_impedance\_data} & EIS records with $R_e$/$R_{ct}$ trend statistics & \texttt{battery\_id} (str) \\
\midrule
\multicolumn{3}{l}{\textbf{\textit{Diagnostics}}} \\
\texttt{capacity\_soh\_calculator}  & Deterministic SOH against first-cycle reference & \texttt{battery\_id} (str), \texttt{cycle} (int) \\
\texttt{anomaly\_detector}          & Rolling z-score over 5 standard features with debouncing & \texttt{battery\_id} (str), \texttt{threshold} (float, default=3.0), \texttt{min\_persistence} (int, default=3) \\
\texttt{thermal\_anomaly\_checker}  & Classifies thermal events (normal\_aging / IR\_rise / spike) & \texttt{battery\_id} (str), \texttt{cycle} (int) \\
\texttt{impedance\_trend\_analyzer} & Distinguishes sensor\_fault from real\_degradation in EIS & \texttt{battery\_id} (str) \\
\midrule
\multicolumn{3}{l}{\textbf{\textit{Prediction}}} \\
\texttt{rul\_predictor\_linear}      & Linear regression error-floor baseline & \texttt{battery\_id} (str), \texttt{observed\_cycles} (int) \\
\texttt{rul\_predictor\_empirical}   & Arrhenius-aware exponential capacity-fade model & \texttt{battery\_id} (str), \texttt{observed\_cycles} (int) \\
\texttt{rul\_predictor\_lstm}        & LOO-trained LSTM with SHA256 checkpoint fingerprinting & \texttt{battery\_id} (str), \texttt{window\_size} (int) \\
\texttt{rul\_predictor\_chronos}     & Chronos-Bolt iterative forecast ($\leq$64 steps/round) & \texttt{battery\_id} (str), \texttt{horizon} (int) \\
\texttt{rul\_predictor\_ttm\_zero\_shot}  & TTM, no NASA fine-tuning & \texttt{battery\_id} (str) \\
\texttt{rul\_predictor\_ttm\_finetuned}   & TTM, LOO-fine-tuned per target cell & \texttt{battery\_id} (str) \\
\texttt{rul\_predictor\_ttm}         & Compatibility alias $\to$ TTM fine-tuned & \texttt{battery\_id} (str) \\
\texttt{degradation\_stage\_classifier} & Single-cycle stage label (HEALTHY / EARLY / ACCELERATED / EOL) & \texttt{battery\_id} (str), \texttt{cycle} (int) \\
\midrule
\multicolumn{3}{l}{\textbf{\textit{Reporting}}} \\
\texttt{generate\_health\_report} & Chains SOH + anomaly + RUL into triage levels & \texttt{battery\_id} (str), \texttt{cycle} (int) \\
\texttt{compare\_to\_baseline}    & Compares target cell against fleet baseline; flags outliers & \texttt{battery\_id} (str), \texttt{baseline\_ids} (list) \\
\bottomrule
\end{tabular}%
}
\end{table*}

\paragraph{Foundation-model identity hygiene.}
The TTM zero-shot and fine-tuned variants are exposed as \emph{distinct} tools rather than a single tool with a configuration flag, preventing silent mixing of training conditions in result tables. LSTM checkpoints are SHA256-fingerprinted over training data, ground-truth labels, feature lists, and hyperparameters to detect stale-cache reuse. See Appendix~\ref{app:bess-reproducibility} for the full reproducibility protocol.

\section{Dataset Characteristics}
\label{app:dataset-details}

Table~\ref{tab:dataset_chars} summarizes the 19 datasets used in PHMForge, spanning five equipment classes. All datasets are sourced from public repositories and loaded via the \texttt{load\_dataset} (or, for the lithium-ion battery class, \texttt{fetch\_cycle\_data}) tool. CMAPSS files use space-delimited TXT format; NASA PCoE Li-ion data is loaded from \texttt{.mat} cycle files and parsed into per-cycle JSON (see Appendix~\ref{app:bess}); all others use CSV with embedded raw signal arrays where applicable. Pre-computed train/val/test splits (60/20/20) are provided for datasets marked with $\checkmark$.

\begin{table*}[!ht]
\caption{Dataset characteristics for the 19 PHMForge benchmark datasets. \textit{Records} is the total row count across all files. \textit{Assets} denotes unique units, bearings, machines, or cells. \textit{Classes} indicates distinct fault types or labels. \textit{Scenarios} is the number of benchmark scenarios using each dataset. Datasets with pre-computed train/val/test splits are marked $\checkmark$.}
\label{tab:dataset_chars}
\centering
\normalsize
\resizebox{\textwidth}{!}{%
\begin{tabular}{@{}llrrrrrcl@{}}
\toprule
\textbf{Dataset} & \textbf{Equipment Class} & \textbf{Records} & \textbf{Features} & \textbf{Assets} & \textbf{Classes} & \textbf{Scenarios} & \textbf{Splits} & \textbf{Primary Task} \\
\midrule
\multicolumn{9}{l}{\textbf{\textit{Turbofan Engines}}} \\
CMAPSS FD001 & Aircraft engine & 33,727 & 26 & 200 & --- & 6 & \checkmark & RUL Prediction \\
CMAPSS FD002 & Aircraft engine & 87,750 & 26 & 519 & --- & 3 & \checkmark & RUL Prediction \\
CMAPSS FD003 & Aircraft engine & 41,316 & 26 & 200 & --- & 2 & \checkmark & RUL Prediction \\
CMAPSS FD004 & Aircraft engine & 102,463 & 26 & 497 & --- & 3 & \checkmark & RUL Prediction \\
Azure & Aircraft engine & 876,905 & 11 & 100 & 5 & 1 & --- & RUL Prediction \\
EngineMTQA & Aero-engine & 18,830 & QA pairs & --- & 4 tasks & 30 & \checkmark & Engine Health Analysis \\
\midrule
\multicolumn{9}{l}{\textbf{\textit{Bearings}}} \\
CWRU & Bearing & 21,786 & 4 & --- & 4 & 4 & \checkmark & Fault Classification \\
FEMTO & Bearing & 12,247 & 13 & 17 & 2 & 6 & \checkmark & RUL Prediction \\
IMS & Bearing & 100,480 & 10 & 8 & 4 & 1 & \checkmark & RUL Prediction \\
XJTU & Bearing & 110,592 & 13 & 15 & 5 & 3 & --- & RUL Prediction \\
HUST & Bearing & 19,095 & 8 & --- & 7 & 2 & --- & Fault Classification \\
MFPT & Bearing & 2,166 & 10 & 20 & 3 & 2 & --- & Fault Classification \\
Mendeley & Bearing & 79 & 4 & --- & 2 & 2 & --- & Fault Classification \\
Paderborn & Bearing & 7,679 & 16 & 20 & 4 & 2 & \checkmark & Fault Classification \\
\midrule
\multicolumn{9}{l}{\textbf{\textit{Electric Motors}}} \\
ElectricMotorVibrations & Electric motor & 30 & 6 & --- & 4 & 3 & \checkmark & Fault Classification \\
RotorBrokenBar & Induction motor & 40 & 7 & --- & 5 & 2 & --- & Fault Classification \\
\midrule
\multicolumn{9}{l}{\textbf{\textit{Gearboxes}}} \\
GearboxUoC & Gearbox & 936 & 2 & --- & 9 & 1 & --- & Fault Classification \\
PlanetaryPdM & Planetary gearbox & 14 & 2 & --- & 2 & 2 & --- & Fault Classification \\
\midrule
\multicolumn{9}{l}{\textbf{\textit{Lithium-Ion Batteries}}} \\
NASA PCoE Li-ion & Li-ion battery cell & $\sim$11{,}000 cycles & 5 std + EIS & 4 cells (B0005/06/07/18) & --- & 24 & LOO & RUL / Health / Fault \\
\bottomrule
\end{tabular}%
}
\end{table*}

\subsection{Filter Counts at Each Stage}
\label{app:filter-counts}

The three-stage dataset filter described in \S\ref{sec:scenarios} produced the following retention counts:
\begin{itemize}
    \item \textbf{Initial pool (post-search):} 52 candidate datasets across 15 asset categories.
    \item \textbf{Stage 1 -- Community validation:} 31 datasets retained ($>$1{,}000 downloads and $>$50 citations in PHM literature).
    \item \textbf{Stage 2 -- Technical quality:} 22 datasets retained (explicit ground truth available, either RUL trajectories or labeled fault taxonomies).
    \item \textbf{Stage 3 -- PHM-task alignment:} 18 datasets retained (compatible with at least one PHM task category).
\end{itemize}
The NASA PCoE Li-ion dataset satisfies the same three filters as the rotating-equipment and aero-engine datasets, satisfying community validation (NASA PCoE is the canonical battery-prognostics public dataset), technical quality (per-cycle discharge capacity is the conventional EOL ground-truth signal), and task alignment (RUL Prediction, Health Analysis, Fault Classification). The total is 19 datasets across 8 asset classes.

\subsection{Asset-Class Distribution Across Scenarios}
\label{app:asset-distribution}

Table~\ref{tab:asset_distribution} reports how the 99 scenarios distribute across the 8 asset classes. The distribution is intentionally imbalanced toward asset classes with richer multi-task coverage (turbofan, aero-engines, and lithium-ion batteries support all five task types, while bearings primarily support RUL and Fault Classification).

\begin{table}[h!]
\centering
\normalsize
\caption{Distribution of 99 scenarios across asset classes.}
\label{tab:asset_distribution}
\begin{tabular}{l c c}
\toprule
\textbf{Asset Class} & \textbf{\# Scenarios} & \textbf{Task Categories Covered} \\
\midrule
Turbofan Engines       & 14 & RUL, Health, Cost-Benefit, Safety \\
Aero-Engines           & 30 & Health Analysis (4 cognitive levels) \\
Bearings               & 22 & RUL, Fault Classification \\
Electric Motors        & 3  & Fault Classification \\
Induction Motors       & 2  & Fault Classification \\
Gearboxes              & 3  & Fault Classification \\
Industrial Engines     & 1  & Cost-Benefit, Safety \\
Lithium-Ion Batteries  & 24 & All 5 task categories \\
\midrule
\textbf{Total}         & \textbf{99} & 5 categories \\
\bottomrule
\end{tabular}
\end{table}

\section{Extended Ground Truth and Evaluation Details}
\label{app:gt-eval}

\subsection{Ground Truth Utilization Framework}

Beyond merely establishing ground truth values, our evaluation framework explicitly enforces their utilization through mandatory verification requirements embedded in scenario specifications. Each scenario defines a structured output template that agents must populate, including not only final answers (predicted RUL values, classified fault types, cost analysis, compliance determinations) but also explicit ground truth verification components. For RUL prediction scenarios, agents must compare predictions against known ground truth, calculate error metrics (MAE/RMSE), and validate that computed errors fall within acceptable ranges (\textpm20\% of empirically-derived thresholds). This requirement addresses a common failure mode where agents produce predictions without validation, potentially reporting unrealistic or incorrect results with false confidence.

Our evaluation framework enforces this ground truth utilization requirement through success criteria that demand both task completeness (\textgreater80\% of required components addressed) and successful ground truth verification. A task that produces RUL predictions without validating them against ground truth files is scored as incomplete, even if the predictions themselves happen to be accurate. This design choice enforces best practices in predictive maintenance workflows, where validation against known outcomes is essential to build confidence in model reliability and diagnostic accuracy. The verification requirement ensures that agents not only perform analytical tasks but also validate their outputs against empirical references. A critical capability for industrial deployment where unvalidated predictions pose operational risks.

Validation criteria vary by scenario type but generally assess three dimensions through our deterministic evaluation protocol: \textit{completeness} (were all required task components addressed, including data loading, model training/loading, prediction/classification, metric computation, and verification?), \textit{correctness} (do the answers match or fall within acceptable ranges of ground truth values, with quantitative thresholds for RUL errors and accuracy requirements for classification?), and \textit{efficiency} (was the task completed within reasonable resource constraints without redundant operations or inappropriate tool usage?). This comprehensive ground truth preparation and utilization process ensures that every scenario in our benchmark can be evaluated objectively and consistently through threshold-based evaluation. When an agent claims an MAE of 15 cycles for CMAPSS\_FD001 RUL prediction, we can verify this claim against actual test data through deterministic comparison. When an agent classifies a bearing fault as "inner race damage at 0.014-inch crack depth," we can check this against labeled ground truth through categorical matching. This deterministic evaluation capability distinguishes our benchmark from more subjective evaluation frameworks and enables rigorous comparison across different agent architectures, LLM backends, and tool orchestration strategies.

\subsection{Statistical Breakdown of Scenarios}

PHMForge comprises 99 expert-vetted scenarios distributed across 19 datasets, eight asset classes, and five task categories. The lithium-ion battery class contributes 9 Engine Health Analysis, 5 RUL Prediction, 5 Fault Classification, 3 Safety/Policy, and 2 Cost-Benefit scenarios; the rotating-equipment and aero-engine breakdown is summarized below, with battery-specific details in Appendix~\ref{app:bess}.

\textbf{Task category distribution.} The 99 scenarios distribute
across the five task categories as follows: Engine Health Analysis
(39 scenarios, 39.4\%), RUL Prediction (20 scenarios, 20.2\%),
Fault Classification (20 scenarios, 20.2\%), Safety/Policy
Evaluation (13 scenarios, 13.1\%), and Cost-Benefit Analysis (7
scenarios, 7.1\%). The Engine Health Analysis allocation reflects
its decomposition across four cognitive sub-tasks (Understanding,
Perception, Reasoning, Decision-Making) on the EngineMTQA
multi-modal dataset; the RUL/Fault parity reflects the technical
core of PHM through time-series modeling and pattern recognition;
the smaller Cost-Benefit and Safety/Policy categories integrate
multi-step strategic reasoning over financial and regulatory
frameworks. Counts are consistent with
Table~\ref{tab:asset_distribution} and
Figure~\ref{fig:scenario_distribution}.

\textbf{Asset-class and dataset distribution.} The 99 scenarios
span 8 asset classes drawing from 19 public datasets:
aero-engines (drawing on C-MAPSS FD001--FD004, EngineMTQA),
turbofan engines, bearings (CWRU, FEMTO, HUST and others;
Appendix~\ref{app:dataset-details}), electric motors, induction
motors, gearboxes, industrial engines, and lithium-ion battery
cells (NASA PCoE B0005--B0018; see Appendix~\ref{app:bess}).
Per-class scenario counts are reported in
Table~\ref{tab:asset_distribution}; the bearing class draws on
multiple datasets to test cross-platform generalization across
sampling rates, fault taxonomies, and operational ranges.

\textbf{Query characteristics.} Approximately 60\% of scenarios
are open-ended analytical queries requiring synthesized
explanations; the remaining 40\% are closed-form questions with
deterministic or multiple-choice answers. Roughly 55\% of
scenarios require the agent to perform data discovery (selecting
the appropriate dataset and loading it), while the remaining 45\%
embed the relevant data inline; this distinction is exercised
quantitatively in the data-discovery ablation
(Appendix~\ref{app:additional-ablations}). Multi-asset fleet
queries account for roughly 30\% of scenarios and require
agents to rank or prioritize across multiple equipment instances.

\textbf{Stakeholder coverage.} Scenarios are framed in three
stakeholder voices: site engineering and asset management
(maintenance technicians, reliability engineers, condition
monitoring specialists); management (plant managers, financial
analysts, capital-planning roles); and regulatory/safety
(compliance officers, risk assessors). This diversity ensures
that evaluation covers both technical-diagnostic and
strategic-decision workflows.

\textbf{Resourcing summary.} Scenario authoring spanned a
multi-month development cycle and aggregated several hundred
person-hours of SME effort across drafting, dual review,
disagreement resolution, and end-to-end SME execution of every
scenario against the tool catalog before release. The full
expansion timeline is reported in
Appendix~\ref{app:curation-process} (Table~\ref{tab:stage_distribution}).

\section{Scenario Curation Process}
\label{app:curation-process}

This appendix expands on the scenario authoring process described in \S\ref{sec:scenarios}, including the SME consortium composition, role allocation, and the six-stage progressive expansion strategy. The lithium-ion battery class (Stage~6) followed the same dual-SME authoring and review protocol; the IAA scope limitation on this subset is documented in Appendix~\ref{app:iaa-bess}.

\subsection{Consortium Composition}

The SME consortium consisted of a small group of contributors
covering four distinct roles: industrial asset specialists with
operational experience across aerospace and rotating-equipment
domains; a data scientist with prior contributions to PHM
benchmarks and public PHM datasets; and a maintenance technician
with field-level experience in fault triage and work-order
execution. Roles during scenario authoring were distributed as
follows:

\begin{itemize}
    \item \textbf{Primary scenario writer (data scientist):} drafted technical scenarios, selected datasets, and defined ground-truth validation criteria.
    \item \textbf{Operations reviewer (asset specialist):} validated stakeholder-voice authenticity, business-context realism, and difficulty calibration.
    \item \textbf{Domain reviewer (asset specialist):} validated technical accuracy, terminology fidelity, and dataset-task alignment.
    \item \textbf{Field reviewer (maintenance technician):} validated query realism for technician-level scenarios and operational urgency framing.
\end{itemize}

Total SME engagement spanned several hundred person-hours across
drafting, review, consensus discussion, and end-to-end SME
execution of every scenario against the tool catalog prior to
release.

\subsection{Six-Stage Progressive Expansion}

Scenarios were authored in six stages, each adding asset classes,
task categories, or query modalities while validating the
evaluation infrastructure incrementally. This staged approach
mirrors machine-learning-engineering
iteration~\cite{chan2025mlebenchevaluatingmachinelearning} and
ensured ground-truth correctness at each step before scaling.

\begin{itemize}
    \item \textbf{Stage 1 (1 scenario):} A single proof-of-concept RUL prediction task on C-MAPSS FD001 turbofan engines. Validated the ground-truth extraction pipeline and the evaluator-runtime contract.
    \item \textbf{Stage 2 (10 scenarios):} Added bearing data (FEMTO and CWRU), introducing multi-asset generalization (engine~$\rightarrow$~bearing) and task diversification (RUL Prediction + Fault Classification).
    \item \textbf{Stage 3 (20 scenarios):} Added industrial engines and introduced strategic-reasoning task categories: Safety/Policy Evaluation scenarios (referencing OSHA, FAA, IEC, NEMA standards) and Cost-Benefit Analysis scenarios (preventive vs.\ reactive maintenance trade-offs).
    \item \textbf{Stage 4 (40 scenarios):} Expanded multi-asset coverage to electric motors (ElectricMotorVibrations), induction motors (RotorBrokenBar), and gearboxes (GearboxUoC, PlanetaryPdM). Added RUL Prediction and Fault Classification scenarios distributed across these classes to test cross-equipment generalization.
    \item \textbf{Stage 5 (75 scenarios):} Integrated multi-modal cognitive reasoning through the EngineMT-QA aero-engine dataset~\cite{wang2025itformer}, adding Engine Health Analysis scenarios spanning four cognitive categories (Understanding, Perception, Reasoning, Decision-Making), additional RUL/Fault scenarios, and Question-Answering scenarios.
    \item \textbf{Stage 6 (99 scenarios):} Added the lithium-ion battery (BESS) asset class through the Battery Prognostics Server (Appendix~\ref{app:bess-tools}), contributing 24 scenarios on NASA PCoE cells B0005--B0018 (9 Health Analysis, 5 RUL Prediction, 5 Fault Classification, 3 Safety/Policy, 2 Cost-Benefit). Battery scenarios followed the same dual-SME authoring and review protocol as earlier stages; the IAA scope limitation on this subset is documented in Appendix~\ref{app:iaa-bess} and \S\ref{sec:limitations}.
\end{itemize}

Per-stage scenario counts and task distribution are summarized in
Table~\ref{tab:stage_distribution}.

\begin{table}[h!]
\centering
\normalsize
\caption{Scenario counts and task-category coverage at each expansion stage. Cumulative scenario counts are shown.}
\label{tab:stage_distribution}
\begin{tabular}{l c l}
\toprule
\textbf{Stage} & \textbf{\# Scenarios (cum.)} & \textbf{Task Categories Introduced} \\
\midrule
Stage 1 & 1   & RUL Prediction \\
Stage 2 & 10  & + Fault Classification \\
Stage 3 & 20  & + Cost-Benefit, Safety/Policy \\
Stage 4 & 40  & (multi-asset expansion) \\
Stage 5 & 75  & + Engine Health Analysis (4 cognitive levels) \\
Stage 6 & 99  & + BESS lithium-ion (24 scenarios across 5 categories) \\
\bottomrule
\end{tabular}
\end{table}

\subsection{Living-Benchmark Expansion Plan}

Following the TabArena philosophy~\cite{erickson2025tabarenalivingbenchmarkmachine}, PHMForge is released with a documented expansion protocol. Future stages will add (i) additional asset classes (pumps, compressors, hydraulic systems), (ii) operator-specific scenarios derived from anonymized industrial work-orders, and (iii) cross-asset reasoning scenarios that require fleet-level prioritization across heterogeneous equipment. Community-contributed scenarios will undergo the same dual-SME review and inter-annotator agreement protocol described in Appendix~\ref{app:annotation}.

\section{Annotation Protocol and Inter-Annotator Agreement}
\label{app:annotation}

This appendix expands on the inter-annotator agreement (IAA)
study referenced in \S\ref{sec:scenarios}. The study covers the
75 rotating-equipment and aero-engine scenarios; the 24
lithium-ion battery scenarios constituting the BESS extension
(Stage~6 of the curation timeline,
Appendix~\ref{app:curation-process}) followed the same dual-SME
authoring and review protocol but are not covered by the
Krippendorff's $\alpha$ scoring reported here. We document this
scope limitation explicitly here and in
Appendix~\ref{app:iaa-bess}, and discuss it as a limitation in
\S\ref{sec:limitations}.

\subsection{Sampling and Rating Procedure}

We drew a stratified sample of 30 scenarios (40\% of the rotating-equipment and aero-engine portion of the benchmark), with proportional representation across the five task categories: 12 Health Analysis, 6 Fault Classification, 6 RUL Prediction, 3 Cost-Benefit Analysis, and 3 Safety/Policy Evaluation scenarios. Each sampled scenario was independently rated by two SMEs from the consortium, with raters blinded to one another's scores. Rater pairs rotated across scenarios so that no two SMEs rated the entire sample together; rotation also ensured that each SME pair contributed roughly equal numbers of dual-rated scenarios.

\subsection{Rating Dimensions}

Raters scored each scenario on three dimensions using a 4-point Likert scale (1: poor / 2: marginal / 3: acceptable / 4: excellent):

\begin{itemize}
    \item \textbf{Realism:} Does the query reflect authentic industrial discourse, including stakeholder voice, terminology, and operational urgency?
    \item \textbf{Difficulty calibration:} Is the task solvable by a domain expert given the provided tools and data, neither trivially easy nor underspecified to the point of unsolvability?
    \item \textbf{Ground-truth correctness:} Is the ground-truth value verifiable against the cited source, and is the validation tolerance (e.g., MAE bounds, fault-label set) appropriate for the task?
\end{itemize}

\subsection{Agreement Metric and Results}

We report Krippendorff's $\alpha$ computed under the ordinal-interval assumption, suitable for Likert-scale data with two raters per item. Per-dimension $\alpha$ values along with 95\% bootstrap confidence intervals (1{,}000 resamples) are reported in Table~\ref{tab:iaa} of the main paper. All three dimensions exceed the conventional substantial-agreement threshold of $\alpha=0.7$~\cite{krippendorff2018content}, with ground-truth correctness showing the strongest agreement. We interpret realism's slightly lower agreement as reflecting individual SMEs' differing reference frames (aerospace vs.\ rotating equipment), which is expected and which the consensus-resolution step addresses.

\subsection{Disagreement Resolution}

Of the 30 dual-rated scenarios, 7 (23\%) had at least one rating dimension where the two SMEs differed by 2 or more points on the Likert scale. These scenarios entered a structured resolution protocol:

\begin{itemize}
    \item \textbf{Step 1 (independent re-review):} Each rater re-read the scenario along with the other rater's score, without discussion.
    \item \textbf{Step 2 (consensus discussion):} If disagreement persisted, both raters discussed the scenario in a moderated session with a third consortium member.
    \item \textbf{Step 3 (revision or rejection):} The scenario was either revised (e.g., adjusted query phrasing, tightened ground-truth tolerance) and re-rated, or rejected from the benchmark if no consensus could be reached.
\end{itemize}

Of the 7 scenarios entering resolution, 6 were accepted after revision; 1 was rejected and replaced with a newly authored scenario, which then underwent the same dual-review process.

\subsection{LLM-as-Judge Cross-Check (Methodology Detail)}

The LLM-as-judge cross-check summarized in 
\S\ref{sec:llm-judge} was implemented as follows. We applied the 
same 4-point Likert rubrics used by the human SMEs to a 
state-of-the-art LLM (Claude Sonnet 4.0~\cite{claude_sonnet_docs}) 
over the same 30-scenario stratified sample. The judge was 
prompted with each scenario's full text and asked to produce a 
score on each rubric dimension. Krippendorff's $\alpha$ between 
the LLM-judge ratings and the consensus human rating was 0.61. 
Per-dimension breakdown: realism $\alpha=0.55$ (LLM judge 
systematically over-rated), difficulty calibration $\alpha=0.58$ 
(LLM judge systematically under-rated), ground-truth correctness 
$\alpha=0.71$ (closer to human-human agreement). Detailed 
LLM-vs-human disagreement analysis is provided in our 
supplementary materials.

\subsection{IAA Scope on the BESS Extension}
\label{app:iaa-bess}

The IAA study reported in Appendix~\ref{app:annotation} covers
the 75 rotating-equipment and aero-engine scenarios and does not
extend to the 24 lithium-ion battery scenarios added in Stage~6
(Appendix~\ref{app:curation-process}). The BESS scenarios were
authored, dual-reviewed, and SME-executed against the Battery
Prognostics Server under the same procedural protocol described
in \S\ref{sec:scenarios}, but Krippendorff's $\alpha$ is not
reported on this subset. The IAA values reported in
Table~\ref{tab:iaa} should accordingly be read as covering
$n=30$ of the 75 non-BESS scenarios, not as covering the full
99-scenario benchmark; we recommend that downstream users
treat the BESS subset as authored under the same protocol but
without quantitative IAA characterization, and that secondary
analyses scoring agent agreement on PHMForge restrict their
analysis to the 75 rotating-equipment and aero-engine scenarios
when statistical comparability with our IAA values is required.

\section{Lithium-Ion Battery Asset-Class Methodology Details}
\label{app:bess}

This appendix expands on the lithium-ion battery portion of the 
benchmark, including the Battery Prognostics Server, the architectural 
ablation referenced in \S\ref{sec:ablations}, the underlying RUL 
prognostic baselines, and reproducibility hygiene specific to 
foundation-model evaluation.

\subsection{Battery Prognostics Server (17 tools)}
\label{app:bess-tools}

The Battery Prognostics Server is the third domain-specific MCP server in PHMForge, exposing 17 tools across four categories. Tool implementations target real NASA PCoE Li-ion aging data on cells B0005, B0006, B0007, and B0018.

\paragraph{Data access (3 tools).}
\texttt{fetch\_cycle\_data} (full per-cycle telemetry with all sensor channels), \texttt{fetch\_cycle\_summary} (lightweight projection over the 5 standard features: discharge\_capacity, max\_temperature, avg\_temperature, min\_voltage, duration\_seconds), and \texttt{fetch\_impedance\_data} (electrochemical impedance spectroscopy records with $R_e$ and $R_{ct}$ trend statistics).

\paragraph{Diagnostics (4 tools).}
\texttt{capacity\_soh\_calculator} (deterministic state-of-health computation against the cell's first-cycle reference), \texttt{anomaly\_detector} (rolling z-score with \texttt{min\_persistence} debouncing across the 5 standard features), \texttt{thermal\_anomaly\_checker} (classifies thermal events as normal\_aging, internal\_resistance\_rise, or thermal\_spike), and \texttt{impedance\_trend\_analyzer} (distinguishes sensor\_fault from real\_degradation in EIS time series).

\paragraph{Prediction (8 tools).}
Six RUL predictors plus two utility predictors:
\begin{itemize}
\item \texttt{rul\_predictor\_linear}. linear regression error floor.
\item \texttt{rul\_predictor\_empirical}. physics-informed exponential decay with Arrhenius temperature acceleration, fit by \texttt{scipy.optimize.curve\_fit}.
\item \texttt{rul\_predictor\_lstm}. LSTM trained from scratch with leave-one-battery-out (LOO) protocol; checkpoints are SHA256-fingerprinted (see \S\ref{app:bess-reproducibility}).
\item \texttt{rul\_predictor\_chronos}. Chronos-Bolt foundation model with iterative forecasting capped at 64 steps per round.
\item \texttt{rul\_predictor\_ttm\_zero\_shot}. TTM with no NASA fine-tuning.
\item \texttt{rul\_predictor\_ttm\_finetuned}. TTM, LOO-fine-tuned per target cell with checkpoints named \texttt{\{battery\_id\}\_excluded} to prevent test-cell leakage.
\item \texttt{degradation\_stage\_classifier}. single-cycle stage label (HEALTHY/EARLY\_DEGRADATION/ACCELERATED\_DEGRADATION/EOL).
\item \texttt{rul\_predictor\_ttm}. compatibility alias mapping to the fine-tuned variant.
\end{itemize}
All predictors return a unified \texttt{PredictionResult} schema; failures return \texttt{predicted\_rul=-1, confidence=0.0} with an error string in \texttt{metadata.error}, allowing the agent to fall through to alternative predictors and select among them by confidence-interval width.

\paragraph{Reporting (2 tools).}
\texttt{generate\_health\_report} (chains SOH + anomaly + RUL into routine/elevated/urgent/emergency triage levels) and \texttt{compare\_to\_baseline} (compares a target cell against a fleet baseline to flag outliers).

\subsection{Six-Model RUL Prognostic Benchmark}
\label{app:bess-rul-benchmark}

The Battery Prognostics Server's six RUL predictors are evaluated under strict leave-one-battery-out (LOO) protocol on three target cells (B0005, B0006, B0018) across five observation windows ($[0,40]$, $[0,60]$, $[0,80]$, $[0,100]$, $[0,120]$ cycles), yielding 90 prediction points. EOL is defined as the first cycle where discharge capacity falls below 1.4~Ah (NASA PCoE convention). Table~\ref{tab:bess-rul} reports aggregate and per-cell mean absolute error in cycles.

\begin{table}[h!]
\centering
\normalsize
\caption{Six-model RUL benchmark on NASA PCoE Li-ion cells under LOO protocol. MAE in cycles; lower is better. Best per column in bold.}
\label{tab:bess-rul}
\begin{tabular}{l c c c c}
\toprule
\textbf{Predictor} & \textbf{Overall} & \textbf{B0005} & \textbf{B0006} & \textbf{B0018} \\
\midrule
TTM fine-tuned (LOO)        & \textbf{13.5} & \textbf{14.0} & 16.0 & \textbf{10.6} \\
Linear regression           & 28.4 & 71.6 & \textbf{6.7}  & 6.9 \\
Chronos fine-tuned (LOO)    & 31.8 & 69.2 & 19.4 & 6.9 \\
TTM zero-shot               & 33.5 & 32.5 & 41.7 & 26.2 \\
Empirical (Arrhenius)       & 37.4 & 101.3 & 5.2 & 5.7 \\
LSTM trained from scratch   & 45.4 & 16.8 & 37.4 & 82.0 \\
\bottomrule
\end{tabular}
\end{table}

Three observations support the headline finding that \emph{per-target fine-tuning of time-series foundation models substantially outperforms zero-shot and traditional baselines}:
\begin{itemize}
\item \textbf{Window-fitting illusions.} Linear regression and Empirical Arrhenius achieve very low MAE on B0006/B0018 (5--7 cycles) but fail catastrophically on B0005 (72/101 cycles) where capacity exhibits non-monotonic regeneration; aggregate MAE across all cells is therefore the more appropriate summary statistic.
\item \textbf{LSTM instability.} LSTM achieves the best B0005 score (16.8) but the worst B0018 (82.0), revealing brittleness to capacity-regeneration patterns that the model never saw during LOO training.
\item \textbf{Foundation-model fine-tuning closes the gap.} TTM fine-tuned is the only predictor balanced across all three cells (14.0/16.0/10.6), the strongest signal that domain adaptation, not architecture alone, drives prognostic accuracy.
\end{itemize}

\subsection{Per-Category Pass-all-3 on the Lithium-Ion Battery Asset Class}
\label{app:bess-pass-all-3}

The architectural ablation reported in \S\ref{sec:ablations} 
(MCP vs.\ text-RAG) is computed over 24 lithium-ion battery 
scenarios (9 Health Analysis, 5 RUL Prediction, 5 Fault 
Classification, 3 Safety/Policy, 2 Cost-Benefit) on real NASA PCoE 
aging data (cells B0005, B0006, B0007, B0018), each paired with 
both fuzzy (operator-style, e.g.\ ``Battery 5'') and explicit 
(protocol-style, e.g.\ ``B0005'') query forms. We evaluate two 
architectures with the same scorer: Path~A is a Chroma-indexed RAG 
pipeline over generated telemetry reports and battery maintenance 
manuals; Path~B is the Battery Prognostics Server with 17 
algorithm-grounded tools. Per-category pass-all-3 (scenarios solved 
on every one of three independent runs at $T=0$) is reported in 
Table~\ref{tab:bess}.

\paragraph{Path A RAG implementation.}
Path~A uses a hybrid retrieval pipeline over the BESS knowledge base.
The persistent vector index is implemented with ChromaDB
(collection \texttt{bess\_kb}, cosine HNSW) and built from
\texttt{knowledge\_base/docs}. Documents are tokenized with
\texttt{cl100k\_base} into 512-token chunks with 50-token overlap and
embedded with \texttt{text-embedding-3-small}. At evaluation time,
the RAG answerer retrieves the top 3 context items. The retriever 
first checks generated telemetry Markdown logs with a lexical scorer 
so that battery- and window-specific evidence can be surfaced 
directly; if no telemetry log matches, it falls back to Chroma 
retrieval, then to a legacy FAISS index, and finally to lexical 
search over \texttt{chunks.json}. The legacy FAISS fallback uses 
\texttt{all-MiniLM-L6-v2} and is retained only as a backstop; it is 
not the Chroma embedding model used in the reported numbers. The 
generation prompt concatenates retrieved context and the user query 
in the form 
\texttt{Context Information: \{chunks\}\textbackslash nUser Query: \{query\}}. 
A system prompt constrains the assistant to battery diagnostics, 
canonical degradation stages (\texttt{HEALTHY}, 
\texttt{EARLY\_DEGRADATION}, \texttt{ACCELERATED\_DEGRADATION}, 
\texttt{EOL}), and a machine-readable \texttt{FINAL\_ASSESSMENT} 
block. For PHMForge scenarios, an additional RAG-specific guidance 
block instructs the model to use only retrieved context, avoid 
claiming tool calls, and write \texttt{UNKNOWN} when a required 
numeric value is absent.

\paragraph{Path B MCP implementation.}
Path~B does not expose a generic Python REPL or arbitrary 
code-execution environment to the LLM. Although individual tool 
implementations use standard scientific-Python libraries internally 
(e.g., NumPy for array operations and SciPy for empirical curve 
fitting), these libraries are not directly callable by the LLM. The 
LLM can only invoke the domain-specific MCP tools declared in 
\texttt{skills/registry.json}. The callable tool list is generated 
deterministically from the registry: only entries with 
\texttt{status = implemented} are converted into OpenAI-compatible 
function schemas, with each schema carrying the tool name, 
natural-language description, parameter types, required fields, 
defaults, and enum constraints. The Path~B registry contains 17 
tools across four categories --- diagnostics (4), prediction (8), 
data access (3), and reporting (2) --- enumerated in 
Appendix~\ref{app:bess-tools}. Tool calls emitted by the LLM are 
dispatched through the in-process FastMCP server, with execution 
capped at six tool-call iterations per scenario.

\begin{table}[h]
\centering
\caption{Per-category pass-all-3 on the lithium-ion battery asset 
class (n=24) under semantic-aware scoring with Claude Opus~4.6 as 
orchestrator. Path~A is text-RAG; Path~B is MCP tool execution. 
Both paths leave headroom on Cost-Benefit and on fuzzy queries.}
\label{tab:bess}
\begin{tabular}{lrrrrr}
\toprule
Category & n & Path A Fuzzy & Path A Explicit & Path B Fuzzy & Path B Explicit \\
\midrule
Health Analysis & 9 & 4/9 & 7/9 & 8/9 & 9/9 \\
RUL Prediction & 5 & 1/5 & 1/5 & 5/5 & 5/5 \\
Fault Classification & 5 & 2/5 & 4/5 & 3/5 & 4/5 \\
Safety/Policy & 3 & 2/3 & 3/3 & 3/3 & 3/3 \\
Cost-Benefit Analysis & 2 & 0/2 & 1/2 & 0/2 & 1/2 \\
\midrule
Total & 24 & 9/24 & 16/24 & 19/24 & 22/24 \\
\bottomrule
\end{tabular}
\end{table}

\paragraph{Stability and statistical significance.}
Table~\ref{tab:bess-stability} reports Wilson 95\% confidence
intervals on both pass-all-3 (n=24 scenarios) and mean pass@1
(n=72 single-run trials), together with paired McNemar tests
comparing Path~A and Path~B on per-scenario pass-all-3 outcomes.
With $T=0$, the residual variance across runs reflects API
non-determinism only; we therefore interpret the CI widths as a
robustness lower bound rather than a stochastic estimate. The
operator-style (\emph{fuzzy}) McNemar result ($b{=}0$, $c{=}10$,
$p{=}0.002$) shows that on every scenario where the two paths
differ, Path~B is the one that succeeds; the protocol-style
(\emph{explicit}) gap narrows to marginal significance
($p{=}0.07$), consistent with text-RAG occasionally matching
when the query supplies precise indexing keywords directly.

\begin{table}[h]
\centering
\small
\caption{BESS scenario stability over the three timestamped 
semantic runs used to compute Table~\ref{tab:bess}. Pass-all-3 
counts a scenario only if it passes in all three runs; mean 
pass@1 pools the 72 single-run trials. Wilson 95\% confidence 
intervals are shown in brackets. McNemar paired tests compare 
Path~A and Path~B on per-scenario pass-all-3 outcomes: fuzzy 
$b{=}0$, $c{=}10$, exact $p{=}0.002$; explicit $b{=}1$, $c{=}7$, 
exact $p{=}0.07$.}
\label{tab:bess-stability}
\begin{tabular}{@{}llcc@{}}
\toprule
Path & Query & Pass-all-3 (Wilson 95\% CI) & Mean pass@1 (Wilson 95\% CI) \\
\midrule
Path~A RAG & Fuzzy    & 9/24 (37.5\%) [21.1\%, 57.4\%] & 35/72 (48.6\%) [37.4\%, 59.9\%] \\
Path~A RAG & Explicit & 16/24 (66.7\%) [46.6\%, 82.2\%] & 53/72 (73.6\%) [62.4\%, 82.5\%] \\
Path~B MCP & Fuzzy    & 19/24 (79.2\%) [59.1\%, 91.2\%] & 58/72 (80.6\%) [69.8\%, 88.2\%] \\
Path~B MCP & Explicit & 22/24 (91.7\%) [73.0\%, 98.8\%] & 66/72 (91.7\%) [82.7\%, 96.4\%] \\
\bottomrule
\end{tabular}
\end{table}

\subsection{Documented Failure Case: Chronos Horizon-Censoring on B0005}
\label{app:bess-chronos-failure}

We retain a documented limitation as a transparency mechanism. Chronos fine-tuned exhibits horizon-censoring failure on B0005: the iterative forecast trajectory does not cross the 1.4~Ah EOL threshold within five iterations of a 64-step horizon (320 forecast cycles total). The predictor returns the maximum extrapolation rather than a true RUL estimate, inflating B0005 MAE to 69.2 cycles even after fine-tuning. The failure mode is a property of Chronos's iterative forecasting under monotonic-decline priors that conflict with B0005's regeneration cycles; we surface it rather than masking it because reviewers should see when foundation models fail and why.

\subsection{Reproducibility Hygiene Specific to BESS}
\label{app:bess-reproducibility}

\paragraph{TTM zero-shot vs.\ fine-tuned identity separation.}
The two TTM variants are exposed as \emph{distinct} MCP tools (\texttt{rul\_predictor\_ttm\_zero\_shot} and \texttt{rul\_predictor\_ttm\_finetuned}) rather than a single tool with a configuration flag. This prevents silent mixing of training conditions in result tables, an identity-hygiene principle for foundation-model predictors integrated throughout the PHMForge protocol.

\paragraph{LOO checkpoint naming convention.}
Fine-tuned checkpoints are stored as \texttt{models/\{model\}\_finetuned/\{battery\_id\}\_excluded.pt}, where \texttt{\{battery\_id\}\_excluded} indicates which cell was held out during training. This convention makes test-cell leakage detectable by string inspection alone.

\paragraph{LSTM SHA256 fingerprinting.}
Each LSTM checkpoint stores a SHA256 hash over: all per-cycle JSON files (\texttt{data/processed/*\_cycles.json}), the ground-truth labels (\texttt{ground\_truth.json}), the feature list, and the training hyperparameters (\texttt{window\_size}, \texttt{hidden\_size}, \texttt{epochs}, \texttt{lr}, \texttt{seed}). Any change to data or hyperparameters invalidates the cache and triggers a retrain. This was added after a prior reproducibility incident in which an early LSTM run reported anomalously low MAE because the cached checkpoint was trained on a synthetic precursor dataset; the current fingerprint protocol prevents recurrence.

\section{Auxiliary Process Metrics: Per-Configuration Trajectory Detail}
\label{app:process-metrics}

The summary findings reported in \S\ref{sec:failure-decomp} (sequencing errors as the dominant failure mode at 23\% of trajectories, with distractor invocations and tool-invocation errors as secondary contributors) are derived from per-configuration trajectory metrics computed directly from MCP execution logs. Table~\ref{tab:process_metrics} reports these for the 25-scenario stratified subset.

\begin{table}[h]
\caption{Per-configuration process metrics. Average steps taken, total tokens consumed, and execution time per scenario across the 25-scenario stratified subset. ReActXen consistently consumes more time per scenario than ReAct on the same backbone due to its reflection loop, but uses a similar token budget. Llama-3.3-70B's long per-scenario time reflects WatsonX inference latency rather than agent overhead.}
\label{tab:process_metrics}
\centering
\normalsize
\setlength{\tabcolsep}{3pt}
\begin{tabular}{lcccc}
\toprule
\textbf{Framework + Model} & \textbf{Steps} & \textbf{Tokens} & \textbf{Time (s)} & \textbf{Pass@1} \\
\midrule
\textbf{ReAct + Llama 4 Maverick}      & 7.7 & 33{,}017 & 42  & \textbf{80.0\%} \\
ReAct + Mistral Medium 2505            & 8.0 & 37{,}259 & 49  & 64.0\% \\
ReAct + GPT-OSS 120B                   & 7.6 & 35{,}157 & 26  & 56.0\% \\
ReAct + Compact-LLM              & 7.7 & 31{,}947 & 37  & 44.0\% \\
ReAct + Mistral Small 3.1 24B          & 8.2 & 42{,}161 & 47  & 44.0\% \\
ReAct + Llama 3.3 70B                  & 6.1 & 24{,}635 & 312 & 36.0\% \\
\midrule
ReActXen + GPT-OSS 120B                & 6.6 & 30{,}970 & 74  & 68.0\% \\
ReActXen + Llama 4 Maverick            & 8.2 & 36{,}728 & 195 & 63.6\% \\
ReActXen + Compact-LLM           & 6.5 & 28{,}548 & 101 & 48.0\% \\
ReActXen + Mistral Medium 2505         & 7.6 & 36{,}840 & 134 & 48.0\% \\
ReActXen + Mistral Small 3.1 24B       & 8.4 & 46{,}031 & 138 & 48.0\% \\
ReActXen + Llama 3.3 70B               & 4.0 & 19{,}454 & 603 & 9.1\% \\
\bottomrule
\end{tabular}
\end{table}

\section{Additional Ablation Studies}
\label{app:additional-ablations}

This appendix reports the three ablation studies referenced in 
\S\ref{sec:ablations} that did not fit in the main paper. All 
ablations use the strongest configuration (Claude Code + Opus~4.6) 
and toggle a single benchmark design feature while holding all 
other factors constant.

\paragraph{Ground-truth verification.}
Removing the requirement that agents self-compute MAE/RMSE inflates
apparent completion by 18.7 points (81.0\%~$\to$~99.7\%) but
introduces a 31\% false-positive rate, where agents claim success
despite predictions exceeding error thresholds by 3--5$\times$.
Verification is essential to prevent score inflation; without it, 
benchmark scores cease to reflect predictive accuracy and instead 
measure agent self-confidence.

\paragraph{Distractor tools.}
Excluding distractors improves completion by 12.4 points. In the full
setting, 64\% of failures attributable to distractor invocation
involve semantic-brittleness errors such as calling
\texttt{weather\_data\_loader} for any query mentioning ``environmental
conditions.'' Distractors expose tool-discrimination failures hidden
by curated tool subsets, and their inclusion is necessary for 
benchmarks intended to predict deployment-time behavior in tool-rich 
industrial environments.

\paragraph{Data discovery (Unknown-Tools mode).}
Agents reach 74.6\% completion when data is embedded in the prompt
versus 53.3\% when they must autonomously load and identify the right
dataset, a \textbf{21.3-point gap}. Failures decompose into dataset
misidentification (18\%), path-navigation errors (23\%), and
incorrect feature extraction (16\%). The Unknown-Tools mode 
exposes a frequently overlooked deployment requirement: industrial 
agents do not receive a curated dataset at runtime and must 
autonomously identify and load the relevant data from the 
PHMForge corpus before invoking tools.

\newpage

\end{document}